\documentclass[lettersize,journal]{IEEEtran}
\pdfoutput=1
\usepackage{amsmath,amsfonts}
\usepackage{algorithmic}
\usepackage{algorithm}
\usepackage{array}
\usepackage[caption=false,font=normalsize,labelfont=sf,textfont=sf]{subfig}
\usepackage{textcomp}
\usepackage{stfloats}
\usepackage{url}
\usepackage{verbatim}
\usepackage{graphicx}
\usepackage{cite}
\usepackage{multirow}
\usepackage{xcolor}
\makeatletter
\let\NAT@parse\undefined
\makeatother
\usepackage{hyperref}  
\begin{document}

\title{DiscrimLoss: A Universal Loss for Hard Samples and Incorrect Samples Discrimination}

\author{Tingting Wu, Xiao Ding, Hao Zhang, Jinglong Gao, Li Du, Bing Qin, Ting Liu
\thanks{Tingting Wu, Xiao Ding, Hao Zhang, Jinglong Gao, Li Du, Bing Qin, Ting Liu are with the Faculty of Computing, Harbin Institute of Technology, Harbin 150000, China (e-mail: ttwu@ir.hit.edu.cn; xding@ir.hit.edu.cn; zhh1000@hit.edu.cn; jlgao@ir.hit.edu.cn; ldu@ir.hit.edu.cn; bqin@ir.hit.edu.cn; tliu@ir.hit.edu.cn).}
\thanks{Corresponding author: Xiao Ding.}
}



\maketitle

\begin{abstract}
Given data with label noise (i.e., incorrect data), deep neural networks would gradually memorize the label noise and impair model performance. To relieve this issue, curriculum learning is proposed to improve model performance and generalization by ordering training samples in a meaningful (e.g., easy to hard) sequence. Previous work takes incorrect samples as generic hard ones without discriminating between hard samples (i.e., hard samples in correct data) and incorrect samples. Indeed, a model should learn from hard samples to promote generalization rather than overfit to incorrect ones.
In this paper, we address this problem by appending a novel loss function \emph{DiscrimLoss}, on top of the existing task loss. Its main effect is to automatically and stably estimate the importance of easy samples and difficult samples (including hard and incorrect samples) at the early stages of training to improve the model performance. Then, during the following stages, DiscrimLoss is dedicated to discriminating between hard and incorrect samples to improve the model generalization. Such a training strategy can be formulated dynamically in a self-supervised manner, effectively mimicking the main principle of curriculum learning. Experiments on image classification, image regression, text sequence regression, and event relation reasoning demonstrate the versatility and effectiveness of our method, particularly in the presence of diversified noise levels.

\end{abstract}


\section{Introduction}
\IEEEPARstart{D}{eep} neural networks (DNNs) have achieved extraordinary success in various tasks, such as commonsense reasoning~\cite{lin2019kagnet,feng2020scalable}, natural language generation~\cite{mei2015talk}, speech recognition~\cite{graves2013speech}, and image classification~\cite{he2016deep}. 
However, as a data-driven method, the success is largely attributed to the large-scale human-annotated datasets. Because obtaining massive and high-quality annotations is extremely labor-consuming and infeasible, many researchers often take an inexpensive and imperfect method as a substitution, such as pattern-based extraction in NLP ~\cite{ijcai2020-502} and automatic labeling in multimedia processing~\cite{wang2014video,yao2017exploiting,chaudhary2019enhancing,lin2020tag}. 
These methods inevitably involve incorrect samples (i.e., samples with label noise), 
but DNNs can easily overfit to noisy labels and lead to poor generalization performance
\cite{zhang2016understanding}.
To improve model performance and generalization, curriculum learning (CL) presents a new idea that easy samples should be learned before hard ones during training \cite{elman1993learning,bengio2009curriculum}. This can be done by estimating the importance of each sample directly during training based on the intuitive observation that easy and hard samples behave differently in terms of their respective loss, and can therefore be discriminated.

\begin{figure}[t]
\centering
\includegraphics[width=0.7\columnwidth]{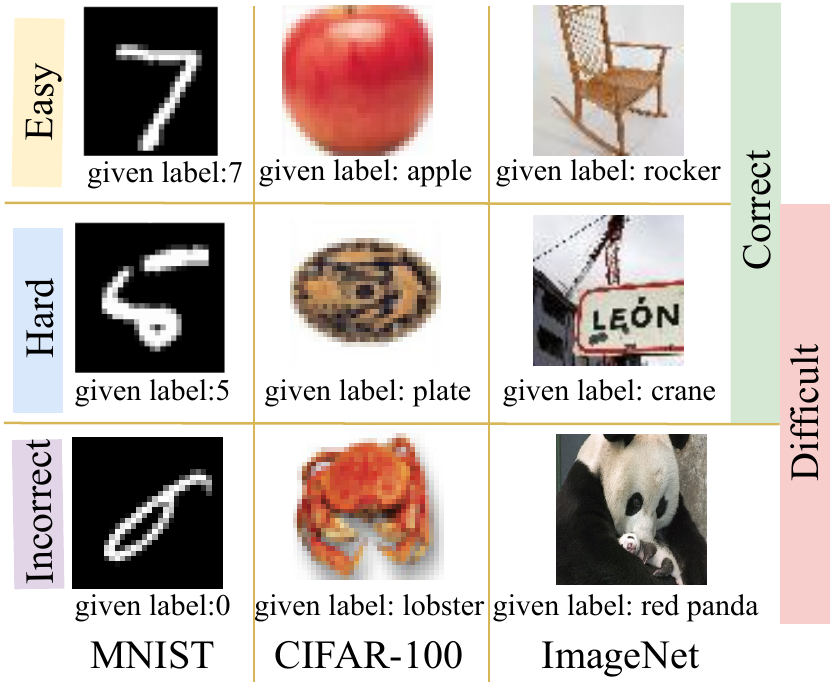}
\caption{An example depicts easy, hard, and incorrect samples for three image datasets. Each sample involves a given label below it as the ground-truth label, which is likely to be noisy.}
\label{example_intro}
\end{figure}
However, previous studies primarily select or reweight samples to suppress the contribution of incorrect samples to the loss function and regard incorrect samples as ordinary hard ones~\cite{kumar2010self,jiang2015self,jiang2018mentornet,saxena2019data,lyu2019curriculum}.
In this situation, hard samples mix with incorrect ones. \cite{toneva2018empirical} state that hard samples rather than incorrect samples are analogs to support vectors, critical for model generalization.

To facilitate this, 
we follow the core principle of curriculum learning and append a novel loss function \textbf{DiscrimLoss} on top of the existing task loss to discriminate between hard samples and incorrect ones.
In contrast to conventional CL, DiscrimLoss focuses on easy, hard, and incorrect samples (exemplified by Fig.~\ref{example_intro}) simultaneously and dedicate to separating hard samples and incorrect samples as much as possible. 
Specifically, DiscrimLoss is a \emph{stage-wise learning strategy}. 
At the early training stages, we focus on improving the model performance by discriminating easy and \emph{difficult samples} (including hard and incorrect samples). 
As we move into the following training stages, we aim at enhancing model generalization by separating hard instances and incorrect ones. In this way, a model can learn from \emph{correct samples} (including easy samples and hard ones) more adequately and avoid memorizing noisy labels.
To this end, we present a stable sample reweight method to automatically estimate the difficulty and importance of samples via an adaptive indicator. 
We learn DiscrimLoss in a self-supervised manner by involving an extra learnable parameter.
Experimental results on six datasets under two modals, comprising benchmark-simulated and real-world label noise datasets, demonstrate consistent gain, which verifies the effectiveness and generality of our method.
\begin{figure}[t]
\centering
\includegraphics[width=0.7\columnwidth]{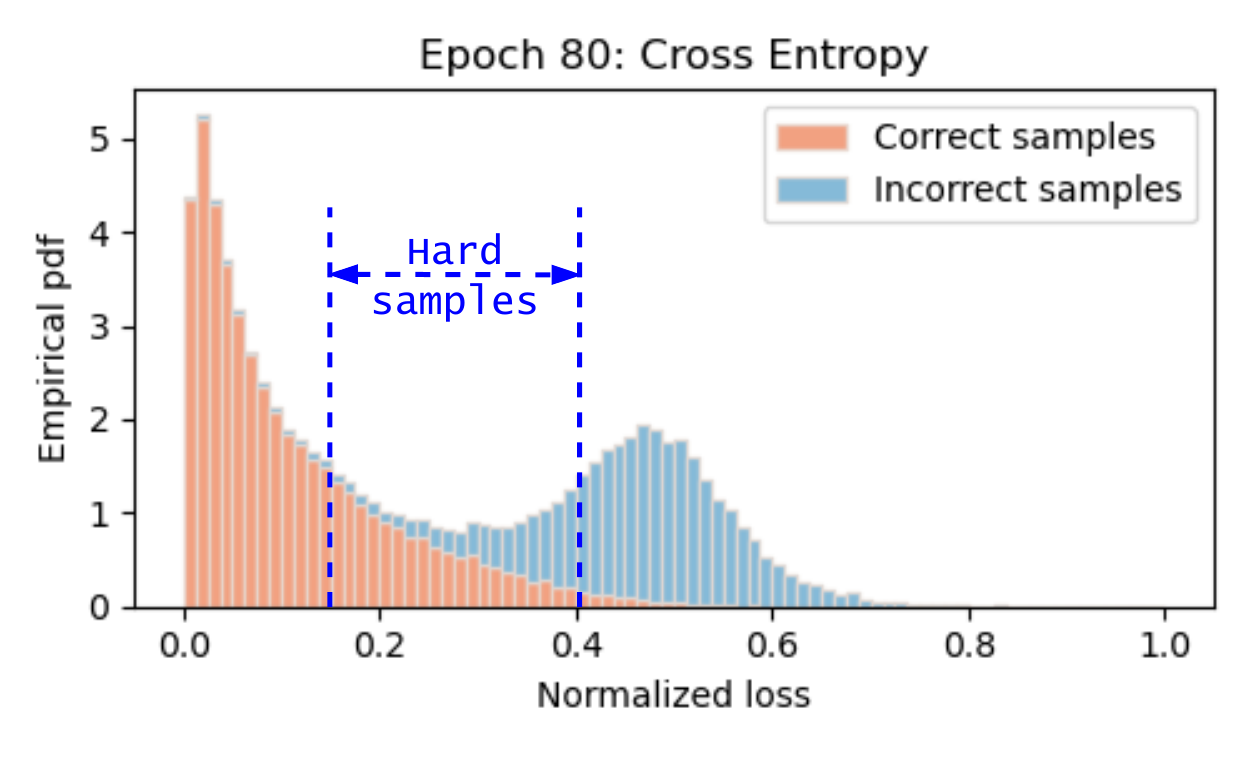}
\caption{Distribution statistics of normalized losses from samples training on CIFAR-100 with 40\% noise with cross-entropy at 80-th epoch.}
\label{Epoch80_ce} 
\end{figure}

\section{DiscrimLoss}\label{Methodology}
We first introduce an essential overview of DiscrimLoss. Then, we present how sample difficulty can be automatically estimated. As a stage-wise learning strategy, sample importance varies with sample difficulty and training stages. DiscrimLoss attempts to harvest a stable weight for each sample to achieve a robust estimation of sample importance. Finally, we demonstrate the complete form of DiscrimLoss and show a detailed analysis of its principle.

\subsection{Overview}\label{Overview}
The basic idea of CL is to estimate a \emph{priori} the difficulty of a given sample by measuring the importance of each sample directly during the training in the form of a weight, such that easy samples with small loss can receive larger weights during the early stages of training.
However, no matter hard and incorrect samples would be learned in previous CL methods, as they cannot explicitly discriminate between these two kinds of samples. Based on the observation of large-scale training samples in Fig.~\ref{Epoch80_ce}, we argue that: (1) the loss of correct samples should be smaller than incorrect ones; (2) partial correct samples, which overlap with incorrect samples, can be regarded as hard samples, whose losses are larger than easy ones and smaller than incorrect ones.
The challenge is how to accurately separate easy, hard and incorrect samples.



\begin{figure}[tbp]
\centering
\includegraphics[width=0.7\columnwidth]{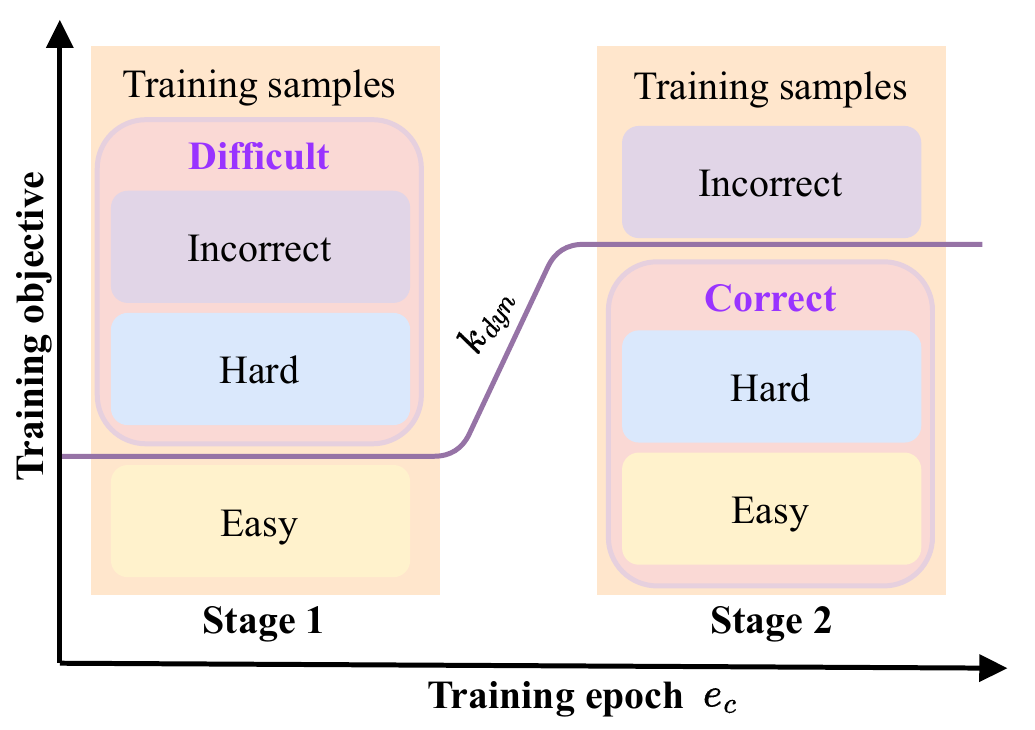} 
\caption{Transfer from Stage 1 to Stage 2 by automatically adjusting an adaptive threshold $k_{dyn}$ in the training process.}
\label{framework} 
\end{figure}

To address this issue, we propose a stage-wise training strategy. As shown in Fig.~\ref{framework}, we first separate easy samples and difficult ones using a threshold $k_{dyn}$, where $k_{dyn} \in \mathbb{R}$ is an adaptive threshold and initialized with a small value at the early training stages (\textbf{Stage~1}). 
The learning capability of the model is weak at this stage.
Most samples are regarded as difficult ones, which would generate large losses.
With the improvement of the model's learning ability, the value of $k_{dyn}$ gradually becomes larger, and more hard samples can be learned by the model.
We use an extra \textbf{learnable parameter} $\delta_{i} \geq 0$ to monitor whether the model can learn knowledge from correct samples, where $\delta_{i}$ denotes the \emph{importance} (or \emph{weight}) of the $i$-th sample. When the model cannot gain any knowledge from correct samples, $k_{dyn}$ reaches its maximum value, so that we can separate hard and incorrect samples using the maximum value of $k_{dyn}$ at this stage (\textbf{Stage~2}).

Formally, let $\left\{\left(\boldsymbol{x}_{i}, y_{i}\right)\right\}_{i=1}^{N}$ denote the data, 
where $\boldsymbol{x}_{i}$ is the $i$-th sample, and $y_{i}$ is its corresponding label.
$l_i=g(M_\theta(\boldsymbol{x}_{i}), y_{i})$ is the loss of sample $\boldsymbol{x}_{i}$ predicted by the model $M$ with trainable parameters $\theta$, where $g(\cdot)$ denotes the \emph{task loss function}. 
We append DiscrimLoss $L_i$ on top of $l_i$ (referred to as \textbf{inner loss} in the following) to discriminate among easy, hard and incorrect samples.
Let $k_{dyn} = k_1 \in \mathbb{R}$ (minimum of $k_{dyn}$) denote the threshold at Stage 1 that preferably distinguishes easy samples and difficult samples according to their respective losses. Similarly, let $k_{dyn} = k_2 \in \mathbb{R}$ (maximum of $k_{dyn}$) be the threshold at Stage 2 that ideally separates hard samples from incorrect ones based on their corresponding losses.

\subsection{Automatic estimation of sample difficulty}\label{modeling on 2 phases}
In the early stages of training, the model focuses on fitting data and improving model performance.
For sample $\boldsymbol{x}_{i}$, if its loss $l_i < k_1$, we take the sample $\boldsymbol{x}_{i}$ as an easy sample and vice versa. In the following stages of training, the model dedicates to enhancing generalization. 
For sample $\boldsymbol{x}_{j}$, if its loss $l_j > k_2$, we take the sample $\boldsymbol{x}_{j}$ as an incorrect sample and vice versa. We take the form $k_{dyn}$ to unify multiple thresholds utilized in varied training stages, and $k_{dyn}$ can be defined as
\small
\begin{equation}
k_{dyn} = \Big(a \cdot \operatorname{Tanh}\big(p(e_{c}-q)\big)+a+1\Big) \cdot k_1,
\end{equation}
\normalsize
where $e_c$ is the current training epoch number, and $\operatorname{Tanh}(\cdot) \in [-1, 1]$ denotes the classical hyperbolic tangent function. 
The purple curve in Fig.~\ref{framework} depicts the relation between $k_{dyn}$ and $e_c$.
$a$, $p$ and $q \in \mathbb{R}$ are hyperparameters. 
$a$ manipulates the switching threshold and makes $k_{dyn}$ fall into the range $[k_1, k_2]$, where $k_2=(1+2a)k_1$. $p$ controls the switching speed, which leads to a more rapid switching when getting smaller and $q$ governs the switching moment. 
\begin{figure}[t] 
\subfloat[ ]{
\includegraphics[width=0.46\columnwidth]{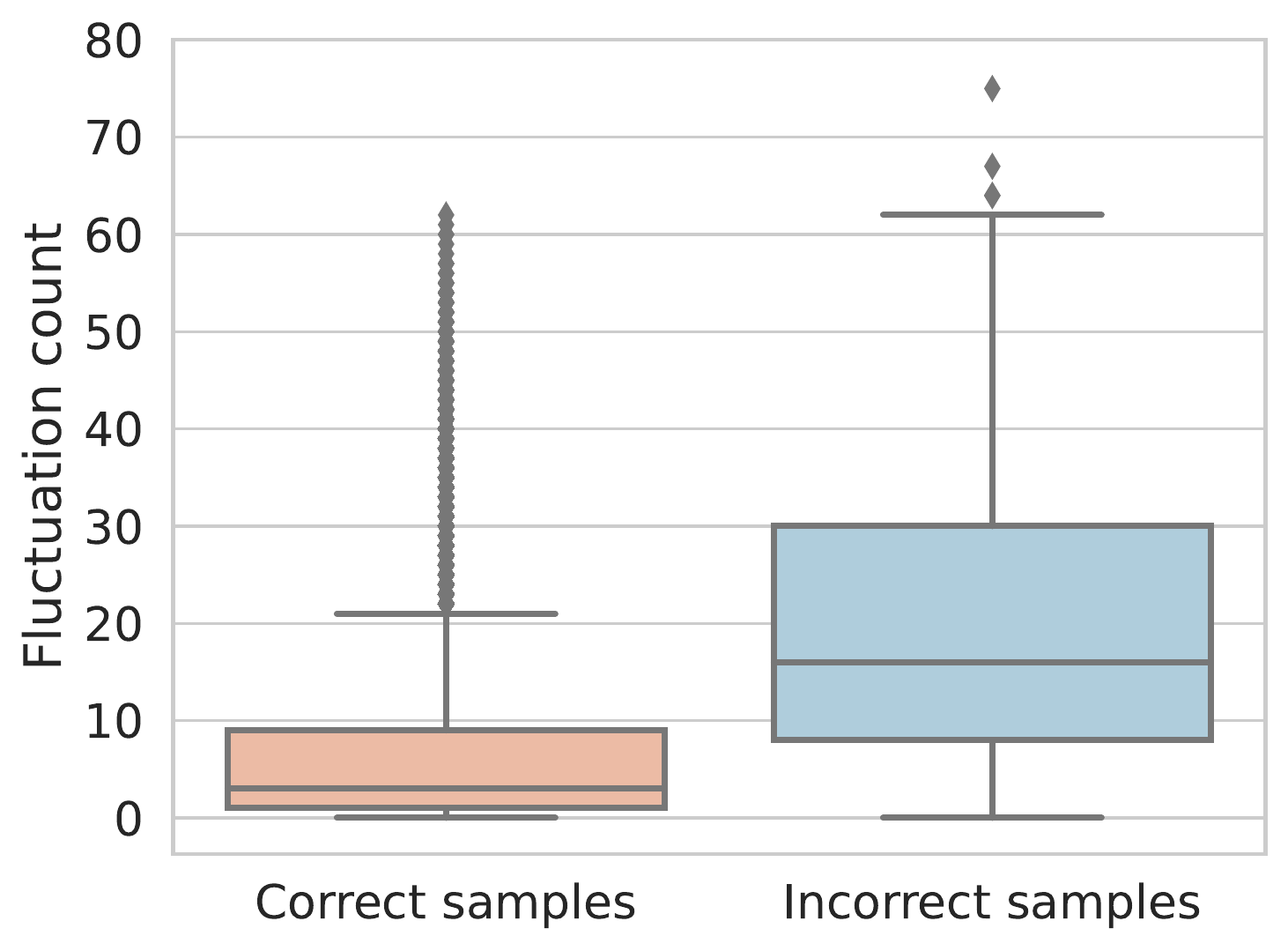}
\label{fluc_statistic} 
}
\subfloat[ ]{
\includegraphics[width=0.48\columnwidth]{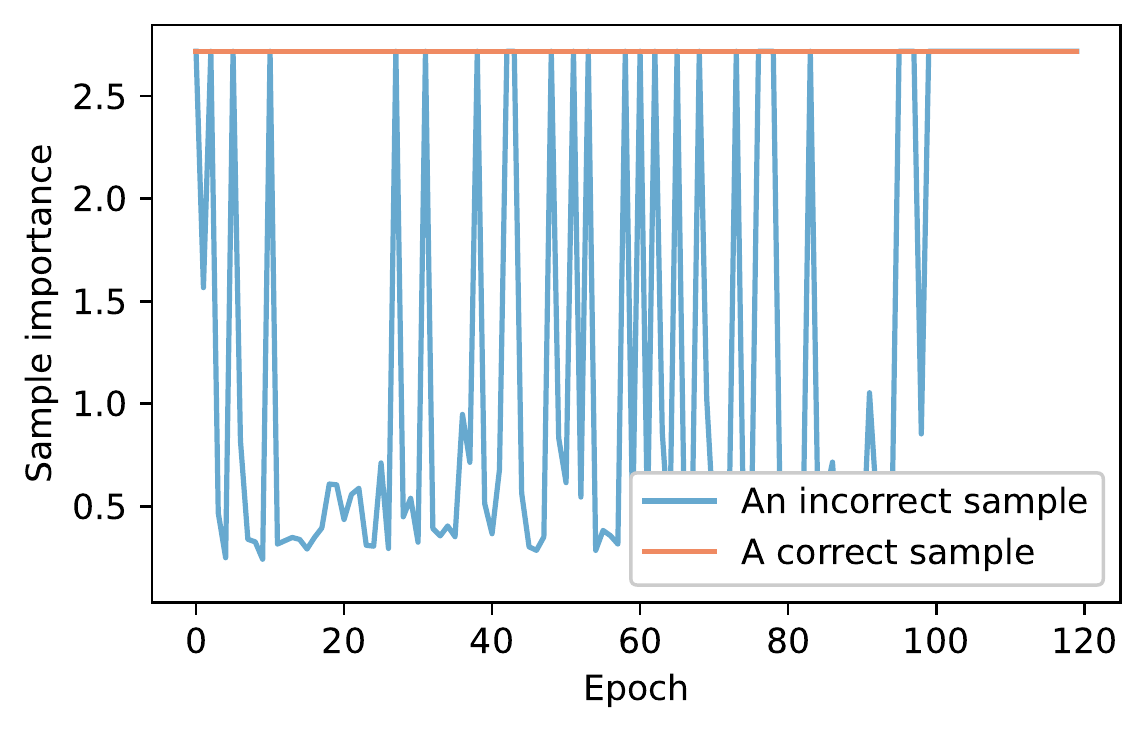}
\label{one_fluc} 
}
\caption{(a) boxplots of fluctuation count\protect\footnotemark in sample importance estimation for correct and incorrect samples. The box represents the region between the lower and upper quartiles of the fluctuation count. The horizontal line in the box represents the median value: \emph{a higher value indicates more volatility}. (b) sample importance estimation of a randomly picked correct/incorrect sample. Both are trained with a recent CL method~\protect\cite{castells2020superloss} on CIFAR-100 with 40\% label noise.}
\label{unstable_case}
\end{figure}
\footnotetext{An estimation will fluctuate when the sample importance estimation of two adjacent epochs for any sample changes more than a specific threshold $T_{fluc}$, where $T_{fluc}=2$ here.}
\subsection{Stable estimation of sample weight}\label{modeling on discrimination}

By estimating sample importance, training samples are arranged in a more meaningful sequence (from easy to hard then to incorrect), which is consistent with the core principle of CL. 
Specifically, at the early stages of training, we increase the importance of easy samples and decrease the importance of difficult ones. For an easy sample, the operation magnifies its contribution to the loss function and obtains a large $L_i$, to accelerate the training process. 
Likewise, at the final stages of training, we increase the importance of hard examples and decrease the importance of incorrect ones. For an incorrect sample, the operation suppresses its contribution to the loss function and gets a small $L_i$, to prevent overfitting.

However, Fig.~\ref{unstable_case}~\subref{fluc_statistic} exhibits incorrect samples are more likely to fluctuate than correct samples when estimating sample importance.
In Fig.~\ref{unstable_case}~\subref{one_fluc}, we randomly picked an incorrect sample, whose importance may be comparable with the correct one at many epochs. In this case, the model may confuse correct and incorrect samples, leading to overfitting. Hence, stable sample importance (or weight) estimation is essential.
As a learnable parameter, the sample weight $\delta_{i}$ is updated by 
\small
\begin{equation}\label{delta_update}
\delta_{i} = \delta_{i} - \tau\frac{\partial L_{i}}{\partial \delta_i}, 
\end{equation}
\normalsize
where $\tau$ denotes the learning rate.
Also, \cite{castells2020superloss} state that $\delta_{i}$ also represents the \textbf{confidence} of the model's prediction. 
From this perspective, we argue that a stable estimation of $\delta_{i}$ should consider the following two components. 
(1) \emph{The initial stage of training would be detrimental to a stable estimation of the sample weight.} 
Random initialization of model parameters in DNNs makes neurons more differentiated, resulting in an unstable weight estimation.
An intuitive approach to make the weight more stable and reliable is to shrink the present gradient in Eq.~(\ref{delta_update}), allowing $\delta_{i}$'s update slower and more cautious and alleviating the impact of model instability.
To this end, we propose a simple but effective constraint called \textbf{Early Suppression} (\textbf{ES}),
which imposes a short-term suppression on the gradient during the early epochs
, and the gradient of $\delta_{i}$ can be defined as
\small
\begin{equation}\label{delta_es}
\frac{\partial L_{i}}{\partial \delta_i} = h(ES(e_c)),\quad
ES(e_c)=
\left\{
\begin{aligned}
&\frac{e_c}{e_s},& &e_{c}<e_{s}, \\ 
&1,& &e_{c} \geq e_{s},
\end{aligned}
\right.
\end{equation}
\normalsize
where $e_{c}$ is the current epoch number. $e_{s}$ denotes the total number of suppressed epochs, which is a hyperparameter. 
$h(\cdot)$ denotes a general form of $\delta_{i}$'s gradient, whose concrete form will be discussed in detail in the next section.

(2) \emph{The losses of hard samples and incorrect ones often fluctuate randomly, which has a bad effect on the stable estimation of the sample weight.}
Specifically, sample importance varies with sample difficulty, and the latter can be characterized by comparing the sample loss with an adaptive threshold $k_{dyn}$. Therefore, $\delta_{i}$'s update depends on the sample loss $l_{i}$. Based on Eq.~(\ref{delta_es}), the gradient of $\delta_{i}$ can be further defined as 
$\frac{\partial L_{i}}{\partial \delta_i} = h(ES(e_c), l_i)$.
However, 
a training example may experience from being identified correctly to incorrectly (or incorrectly to correctly) throughout learning in one task \cite{toneva2018empirical}, which leads to an inconsistent prediction of a single sample. 
This phenomenon is more common in hard samples. To alleviate the random fluctuation of the prediction confidence caused by only one prediction, we incorporate the historical loss into the model loss, i.e., \textbf{Historical Loss} (\textbf{HL}). Then, we have 
\small
\begin{equation}
\frac{\partial L_{i}}{\partial \delta_i} = h(ES(e_c), Avg(l_i)),
\end{equation}
\normalsize
where $Avg(l_i)$ denotes the exponential moving average of $l_i$ with a fixed smoothing parameter $\rho$. 
As to easy samples, model predictions in different epochs tend to be consistent. Using the historical loss does not impair the performance of easy samples. 
Moreover, when we increase the sample importance (for the sample whose $l_i < k_{dyn}$) in Eq.~(\ref{delta_update}), its gradient $\frac{\partial L_{i}}{\partial \delta_i}$ should be negative or vice versa. 
In this case, we use the form $l_i-k_{dyn}$ as another key constraint in the gradient of $\delta_i$, namely
\small
\begin{equation}
\frac{\partial L_{i}}{\partial \delta_i} = h(ES(e_c), Avg(l_i), l_i-k_{dyn}).
\end{equation}
\normalsize

Notably, although sample weights are introduced as learnable parameters, DiscrimLoss is involved only during training. At inference, we take the inner loss as an alternative and hence has no effect on model complexity.

\begin{table}
\setlength\tabcolsep{2pt}
\centering
\caption{The influence on sample weight $\delta_{i}$ and model training with different sample difficulties and training stages.}
\begin{tabular}{|c|c|c|c|c|c|}
\hline
\multirow{2}{*}{Stage} &
\multirow{2}{*}{\large$\frac{\partial L_{i}}{\partial \delta_i}$}                                                                & \multicolumn{2}{c|}{\multirow{2}{*}{\begin{tabular}[c]{@{}c@{}}Sample\\ difficulty\end{tabular}}} & \multirow{2}{*}{\begin{tabular}[c]{@{}c@{}}$\delta_i$'s \\ update\end{tabular}} & \multirow{2}{*}{\begin{tabular}[c]{@{}c@{}}Gradient \\ change\end{tabular}} \\
&                                                                                             & \multicolumn{2}{c|}{}                                                                             &                                                                                &                           \\ \hline
\multirow{3}{*}{1}                                                        & \multirow{3}{*}{\large$\frac{ES(e_c)\left(k_{1}-Avg\left(l_{i}\right)\right)}{\delta_{i}^2}$} & $Avg\left(l_{i}\right) < k_{1}$                                 & easy                     & smaller                                                                        & larger                    \\ \cline{3-6} 
&                                                                                             & \multirow{2}{*}{$Avg\left(l_{i}\right) > k_{1}$}                & hard               & \multirow{2}{*}{larger}                                                        & \multirow{2}{*}{smaller}  \\ \cline{4-4}
&                                                                                             &                                                                       & incorrect                    &                                                                                &                           \\ \hline
\multirow{3}{*}{2}                                                        & \multirow{3}{*}{\large$\frac{ES(e_c)\left(k_{2}-Avg\left(l_{i}\right)\right)}{\delta_{i}^2}$} & \multirow{2}{*}{$Avg\left(l_{i}\right) < k_{2}$}                & easy                     & \multirow{2}{*}{smaller}                                                       & \multirow{2}{*}{larger}   \\ \cline{4-4}
&                                                                                             &                                                                       & hard               &                                                                                &                           \\ \cline{3-6} 
&                                                                                             & $Avg\left(l_{i}\right) > k_{2}$                                 & incorrect                    & larger                                                                         & smaller                   \\ \hline
\end{tabular}
\label{influence_on_sample_weight}
\end{table}
\subsection{Stage-wise loss function}\label{discrimloss form}
Based on the above analysis, we propose a simple possible formulation of DiscrimLoss that meets design for $h(\cdot)$, i.e.,
\small
\begin{equation}\label{complete_form_DiscrimLoss}
L_i = \frac{ES(e_c)\left(Avg\left(l_{i}\right)-k_{dyn}\right)}{\delta_{i}}+\lambda\left(\log \delta_{i}\right)^{2},
\end{equation}
\normalsize
where $\left(\log \delta_{i}\right)^{2}$
denotes a regularization term controlled by the hyperparameter $\lambda>0$.
To better illustrate the underlying principles of DiscrimLoss, we start from the optimization of learnable parameters (i.e., sample weight $\delta_i$ and model parameters $\theta$).
Without considering the regularization term,  the gradient of the loss w.r.t the sample weight $\delta_i$ is given by:
\small
\begin{equation}\label{gradient_delta}
\frac{\partial L_{i}}{\partial \delta_i} = \frac{ES(e_c)\left(k_{dyn}-Avg\left(l_{i}\right)\right)}{\delta_{i}^{2}}.
\end{equation}
\normalsize
Both ES and HL components impose further constraints on $\delta_{i}$'s gradient and make $\delta_{i}$ more stable. 
Based on Eq.~(\ref{gradient_delta}), we summarize the influence on sample weight and model training with different sample difficulties and training stages, as shown in Table~\ref{influence_on_sample_weight}. 
We take a hard sample in the early stages of training (i.e., training stage 1, $k_{dyn}=k_{1}$) as an example. 
As a hard sample (i.e., $Avg\left(l_{i}\right) > k_{1}$), it generates a negative gradient in Eq.~(\ref{gradient_delta}), which leads to the increasing of $\delta_i$ in Eq.~(\ref{delta_update}), i.e., ``larger'' in Table~\ref{influence_on_sample_weight}. 
A larger $\delta_i$ causes a smaller $L_{i}$ in Eq.~(\ref{complete_form_DiscrimLoss}), which suppresses the loss contribution made by a hard sample. Also, a larger $\delta_i$ makes the related parameters update slower. Similarly, the gradient of the loss with respect to the model parameters $\theta$ is given by: 
\small
\begin{equation}\label{gradient_model_parameters}
\frac{\partial L_{i}}{\partial \theta}=\frac{\partial L_{i}}{\partial l_{i}} \cdot \frac{\partial l_{i}}{\partial \theta}=\frac{ES(e_c)}{\delta_{i}} \cdot \frac{\partial l_{i}}{\partial \theta},
\end{equation}
\normalsize
where $\frac{\partial l_{i}}{\partial \theta}$ is the inner loss gradient. 
The historical loss is a constant in Eq.~(\ref{complete_form_DiscrimLoss}), and thus the derivative with respect to $\theta$ is 0, which does not affect the gradient in Eq.~(\ref{gradient_model_parameters}). 
In addition, as $ES(e_c)$ is a constant at each epoch, $\frac{ES(e_c)}{\delta_{i}}$ can be regarded as the sample weight, which is ultra-low in the initial stage of training. At this time, the model lacks reliability, and the DiscrimLoss $L_{i}$ can decouple from the original loss $l_i$. When confidence is getting larger, the DiscrimLoss $L_{i}$ has a positive correlation with the inner loss $l_i$.

\section{Experiments}\label{experiments}

We conduct experiments on four different tasks across CV and NLP to test the effectiveness and generality of DiscrimLoss (Section~\ref{img_classification} to \ref{event_rel_reasoning}). 
We further validate the effect of our method on model generalization performance (Section~\ref{gen_performance_ana}).
Finally, we investigate the influence of each component of DiscrimLoss through an ablation study (Section~\ref{ab_study}).

\subsection{Experiment settings}\label{appendix_experiment_settings}
Before formally introducing the experiments, we first introduce the basic settings of the experiments.
Without loss of generality, we mainly focus on both the classification and regression tasks. For the classification tasks, we take the cross-entropy as the inner loss, and utilize accuracy as the evaluation metric. For the regression tasks, the Mean-Square-Error (MSE) loss ($l_2$) or the smooth-L1 loss (smooth-$l_1$) are taken as the inner loss, and the mean absolute error (MAE) is adopted as the evaluation metric.

We conduct experiments on two real-world noisy datasets and five manually corrupted datasets, where the latter enables us to examine the specific efficacy of DiscrimLoss under different proportions of noise.
Following previous works \cite{saxena2019data,castells2020superloss}, 
we manually corrupt the dataset by introducing symmetric label noise, 
as it can be more challenging to avoid incurring such noise into the model during training \cite{patrini2017making}. To this end, we randomly replace a proportion of the train labels with the other labels drawn from a uniform distribution.

In the training process, we empirically initialize $k_1$ as the global average of the inner loss (simplified as $k_1$ = $\mathrm{GA}$), exponential moving average of the inner loss with a fixed smoothing parameter $\rho'=0.9$ (simplified as $k_1$ = EMA), or a constant $\eta$ assigned via prior knowledge about the specific task (i.e., $k_1=\eta$). For $\rho$ in $Avg(l_i)$, $\rho=0.9$. 

For all results in this paper, methods are implemented with PyTorch and trained on GeForce RTX 2080Ti GPUs. 
We report the mean and standard deviation over five runs. 
All the sample weights are initialized as 1 and are optimized using the standard stochastic gradient descent (SGD). A sample weight $\delta_i$ will be updated when and only when it is present in a mini-batch.
We also refine the hyperparameters for each noise level for datasets with manually corrupted labels. Entire optimal hyperparameter settings are provided in Table~\ref{optimal-hyper-with-noise-rate} and Table~\ref{optimal-hyper} of the Appendix~\ref{appendix_hypersetting}.
Hyperparameters are optimized by a common hyperparameter optimization tool,  \emph{Hyperopt}~\cite{bergstra2013making}.

\begin{figure*}[htbp]
\centering
\subfloat[ ]{
\includegraphics[width=0.50\columnwidth]{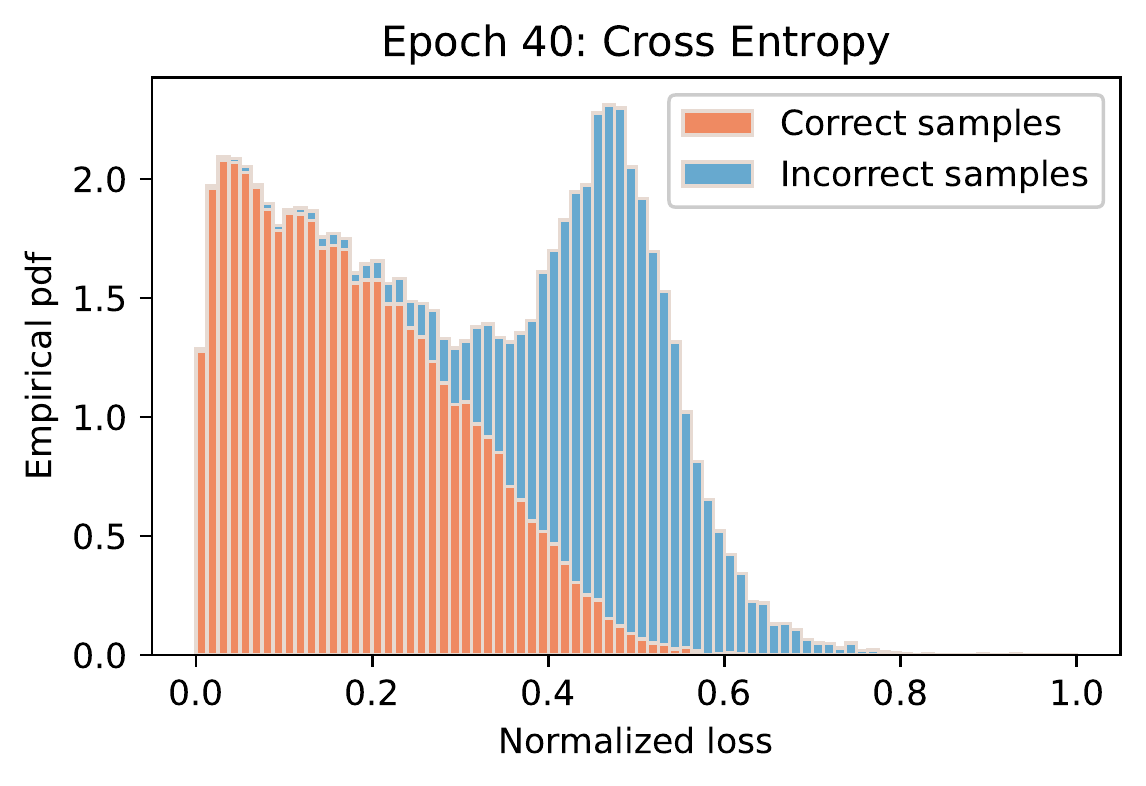}
\label{ce_40} 
}
\hfil
\subfloat[ ]{
\includegraphics[width=0.50\columnwidth]{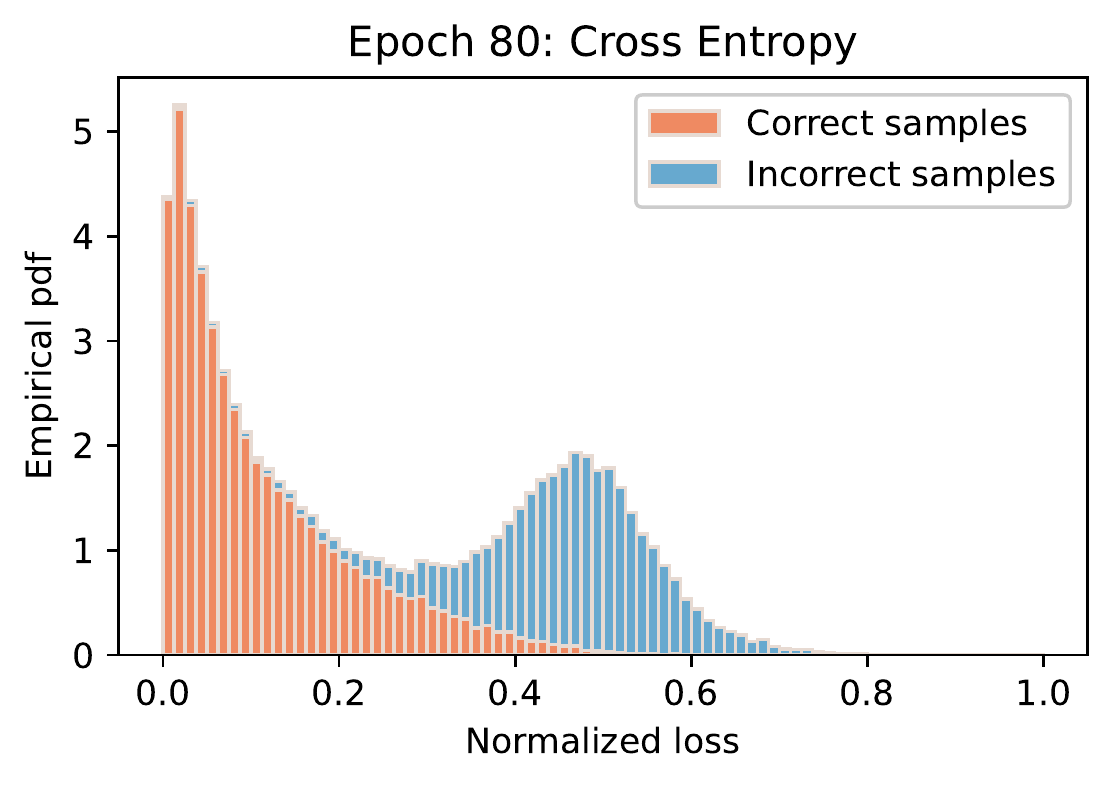}
\label{ce_80} 
}
\hfil
\subfloat[ ]{
\includegraphics[width=0.50\columnwidth]{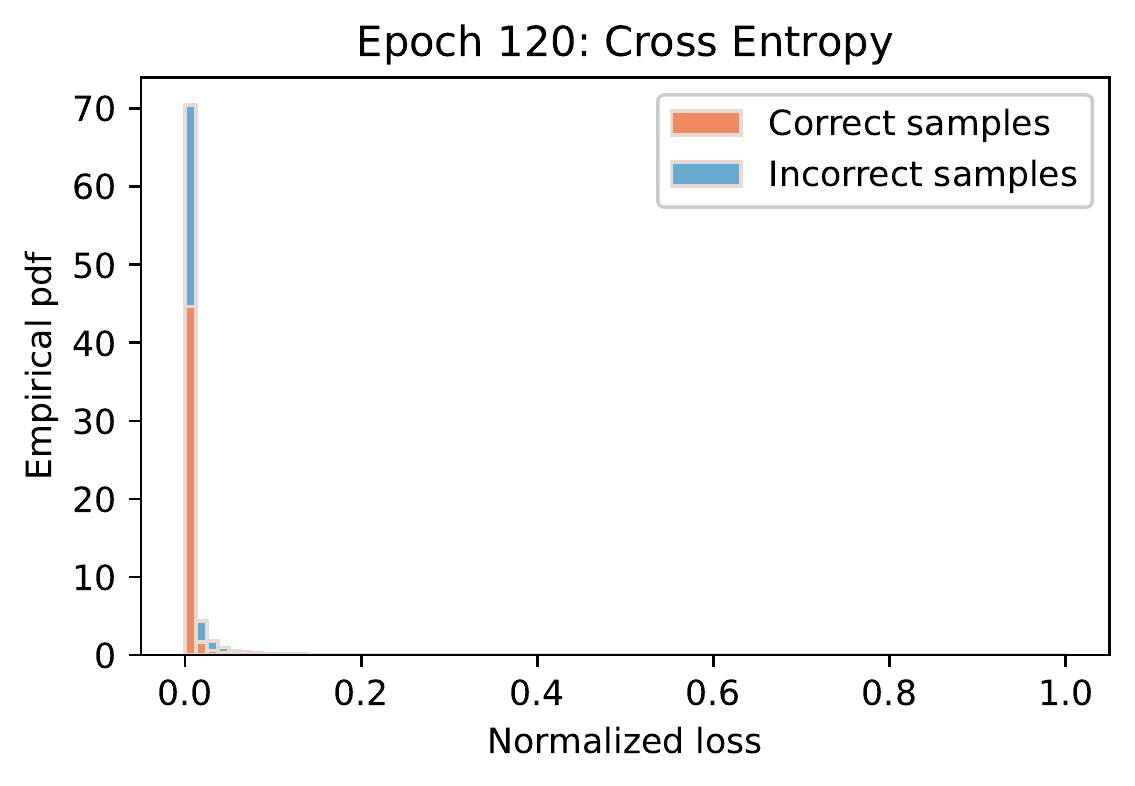}
\label{ce_120} 
}
\hfil
\subfloat[ ]{
\includegraphics[width=0.50\columnwidth]{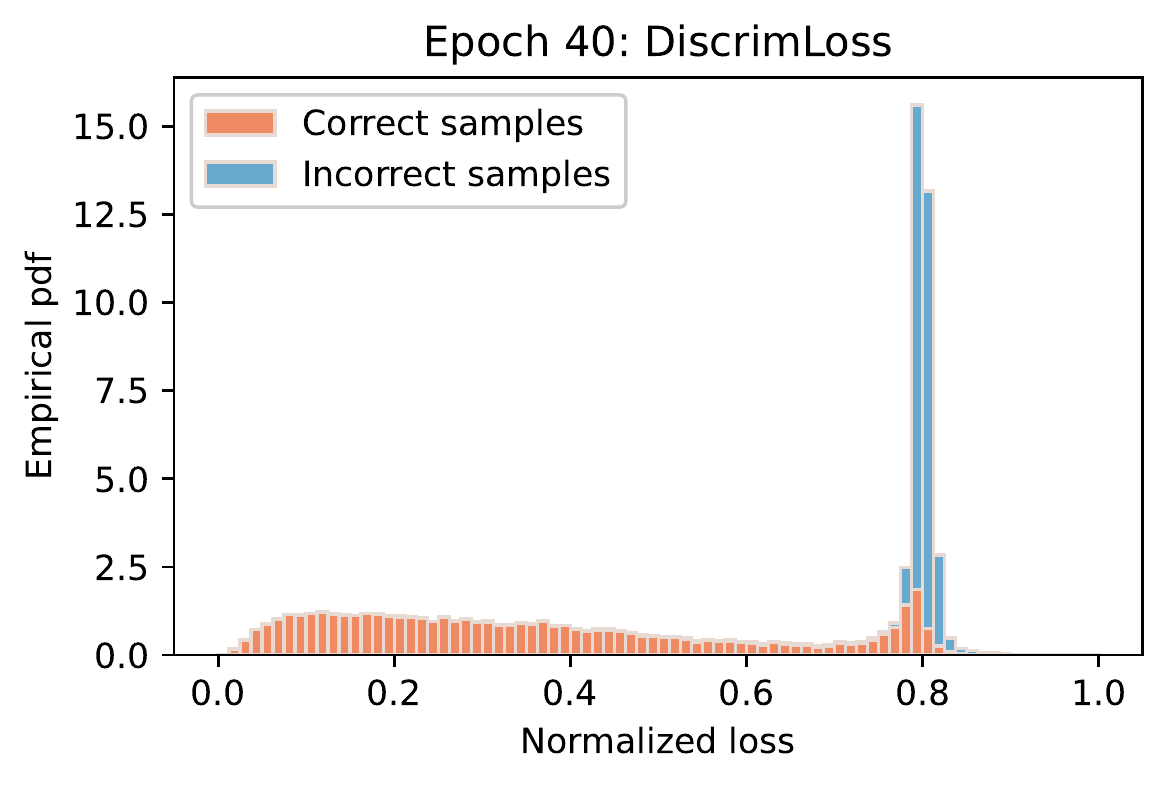}
\label{dl_40} 
}
\hfil
\subfloat[ ]{
\includegraphics[width=0.50\columnwidth]{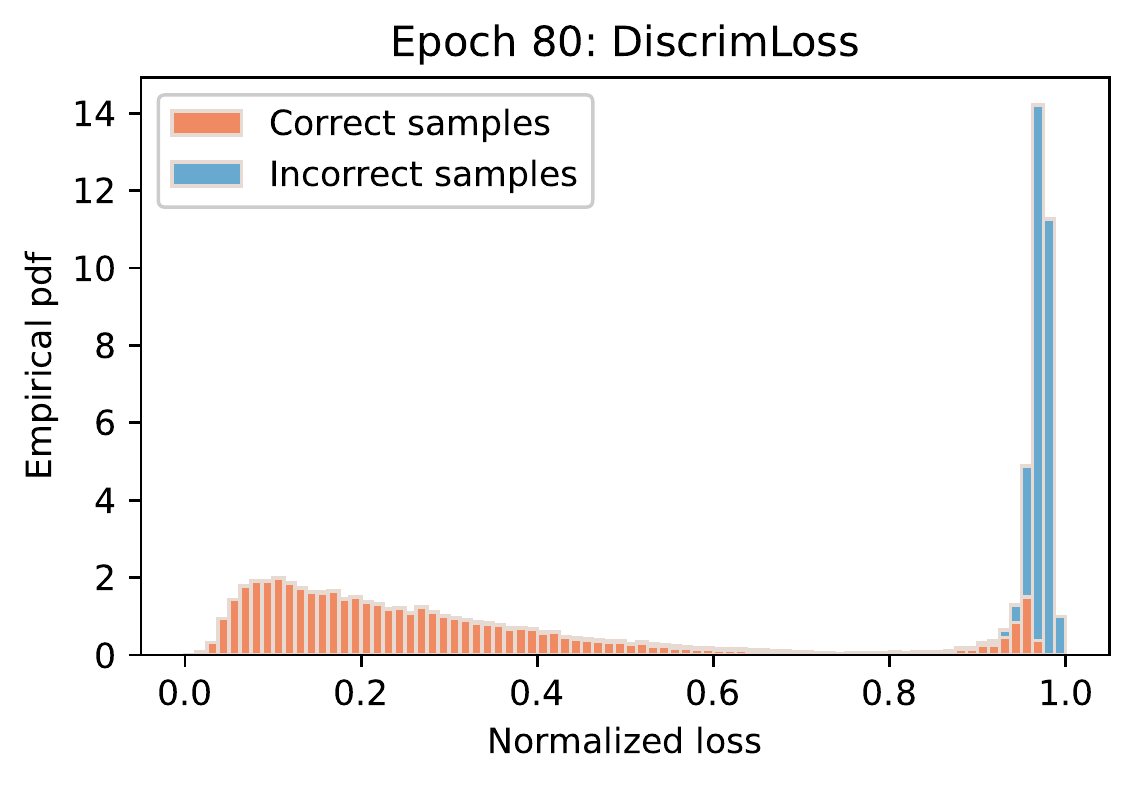}
\label{dl_80} 
}
\hfil
\subfloat[ ]{
\includegraphics[width=0.50\columnwidth]{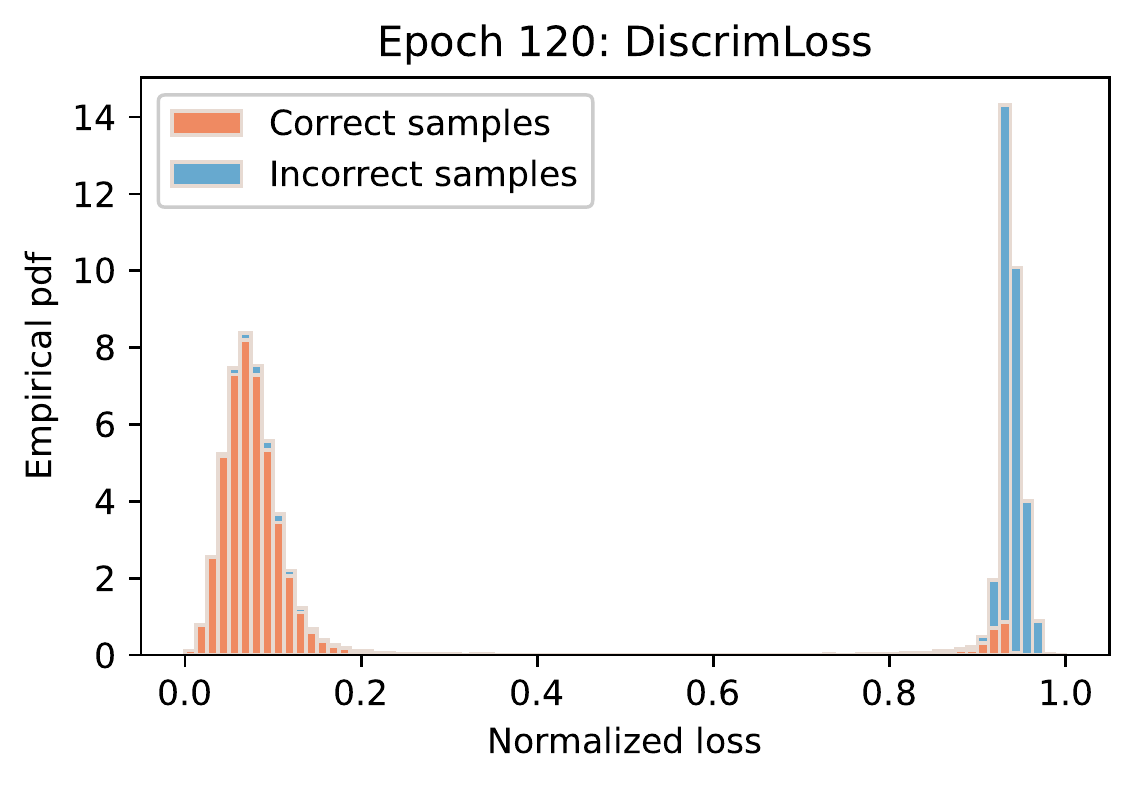}
\label{dl_120} 
}
\caption{Distribution statistics of normalized losses from samples training on CIFAR-100 with 40\% noise. (a) Training with cross-entropy at 40-th epoch. (b) Training with cross-entropy at 80-th epoch. (c) Training with cross-entropy at 120-th epoch. (d) Training with DiscrimLoss at 40-th epoch. (e) Training with DiscrimLoss at 80-th epoch. (f) Training with DiscrimLoss at 120-th epoch.} 
\label{hist_9}
\end{figure*}
\begin{table}[]
\centering
\caption{Statistics of datasets for image classification.}
\begin{tabular}{|c|c|c|c|c|c|}
\hline
\multirow{2}{*}{} & \multirow{2}{*}{\begin{tabular}[c]{@{}c@{}}Train \\ Size\end{tabular}} & \multirow{2}{*}{\begin{tabular}[c]{@{}c@{}}Validation \\ Size\end{tabular}} & \multirow{2}{*}{\begin{tabular}[c]{@{}c@{}}Test \\ Size\end{tabular}} & \multirow{2}{*}{\begin{tabular}[c]{@{}c@{}}Classes \\ \end{tabular}} & \multirow{2}{*}{\begin{tabular}[c]{@{}c@{}}Image \\ Size\end{tabular}} \\
&                                                                            &                                                                              &                                                                           &                                                                         &                                                                        \\ \hline
MNIST     & 60,000  & -      & 10,000       & 10         & 28$\times$28 \\ 
CIFAR-10  & 50,000  & -      & 10,000       & 10         & 32$\times$32 \\ 
CIFAR-100 & 50,000  & -      & 10,000       & 100        & 32$\times$32 \\ 
Clothing1M	& 1,000,000 & 14,313	& 10,526	& 14	& 224$\times$224 \\ \hline
\end{tabular}
\label{datasets_img_classification}
\end{table}

\begin{table}[]
\setlength\tabcolsep{1.0pt}
\centering
\caption{\textcolor{black}{Test accuracy on MNIST under different proportions of label noise.}}
\begin{tabular}{|c|c|c|c|c|c|}%
\hline
\multirow{2}{*}{Method}         & \multicolumn{5}{c|}{MNIST}                                             \\ \cline{2-6}
& 0\%            & 20\%           & 40\%           & 60\%           & 80\%           \\ \hline
MentorNet & -              & 97.21$\pm$0.13 & 93.96$\pm$0.76 & -              & -              \\
Co-teaching & -              & 97.22$\pm$0.18 & 94.64$\pm$0.33 & -              & -              \\
Co-teaching$+$ & -              & 98.11$\pm$0.07 & 95.87$\pm$0.27 & -              & -              \\
JoCoR & -	& 98.06$\pm$0.04	& -	& - & 84.89$\pm$4.55	\\
SuperLoss  & 98.97$\pm$0.06 & 98.87$\pm$0.05 & 98.40$\pm$0.16 & 97.69$\pm$0.14 & 95.53$\pm$0.18 \\ \hline
\multicolumn{1}{|c|}{DiscrimLoss} & \textbf{99.08$\pm$0.03} & \textbf{98.99$\pm$0.01} & \textbf{98.64$\pm$0.01} & \textbf{98.11$\pm$0.09} & \textbf{96.15$\pm$0.16} \\ \hline
\end{tabular}
\label{MNIST-results}
\vspace{-0.4cm}
\end{table}

\begin{table*}[]
\centering
\caption{\textcolor{black}{Test accuracy on CIFAR-10 and CIFAR-100 under different proportions of label noise.}}
\begin{tabular}{|c|c|c|c|c|c|c|c|c|}
\hline
\multirow{2}{*}{Method} & \multicolumn{4}{c|}{CIFAR-10}  & \multicolumn{4}{c|}{CIFAR-100}                                                                         \\ \cline{2-5} \cline{6-9}
& 0\%                     & 20\%                    & 40\%                     & 60\%                    & 0\%                     & 20\%                    & 40\%                    & 60\%                    \\ \cline{1-5} \cline{6-9}
MentorNet               & -                       & 83.26$\pm$0.72          & 78.37$\pm$1.73           & -                       & -                       & 57.27$\pm$1.32          & 49.01$\pm$2.09          & -                       \\
Co-teaching             & -                       & 88.20$\pm$0.27          & 84.45$\pm$0.68           & -                       & -                       & 61.47$\pm$0.41          & 53.44$\pm$0.40          & -                       \\
Co-teaching$+$          & -                       & 86.47$\pm$0.92          & 78.93$\pm$0.74           & -                       & -                       & 64.13$\pm$0.32          & 55.92$\pm$0.81          & -                       \\
Data Parameters         & -                       & -                       & 91.10$\pm$0.70           & -                       & -                       & -                       & 70.93$\pm$0.15          & -                       \\
CurriculumLoss          & -                       & 89.49                   & 83.24                    & 66.2                    & -                       & 64.88                   & 56.34                   & 44.49                   \\
SuperLoss               & 94.97$\pm$0.03          & 92.98$\pm$0.11          & 91.06$\pm$0.23           & 85.48$\pm$0.13          & 78.51$\pm$0.10          & 74.34$\pm$0.26          & 70.96$\pm$0.24          & 62.39$\pm$0.17          \\
SLN                     & -                       & -                       & 80.00$\pm$0.61           & -                       & -                       & -                       & 50.24$\pm$0.41          & -                       \\
CDR                     & -                       & 90.26$\pm$0.31          & 87.19$\pm$0.43           & -                       & -                       & 68.68$\pm$0.33          & 62.72$\pm$0.38          & -                       \\ \cline{1-5} \cline{6-9}
DiscrimLoss             & \textbf{95.99$\pm$0.17} & \textbf{93.67$\pm$0.05} & \textbf{92.07$\pm$0.18} & \textbf{86.82$\pm$0.06} & \textbf{80.77$\pm$0.12} & \textbf{75.06$\pm$0.06} & \textbf{71.04$\pm$0.07} & \textbf{62.46$\pm$0.06} \\ \hline
\end{tabular}
\label{CIFAR10-CIFAR100-results}
\end{table*}

\begin{table}[t]
\centering
\caption{\textcolor{black}{Test accuracy on Clothing1M. Top-2 results are in bold.}}
\begin{tabular}{|c|c|}
\hline
Method                & Test accuracy   \\ \hline
CE                    & 71.12$\pm$0.32 \\
Forward & 71.28$\pm$0.27	\\
Backward & 71.03$\pm$0.33 \\
Co-Teaching           & 72.14$\pm$0.28 \\
JoCoR & 72.30$\pm$0.34 \\
SuperLoss             & 71.89$\pm$0.44  \\ 
SLN                   & 72.95$\pm$0.31 \\ \hline
DiscrimLoss           & \textbf{73.10$\pm$0.11}  \\
SuperLoss-DiscrimLoss & \textbf{73.79$\pm$0.25}  \\ \hline
\end{tabular}
\label{Clothin1M-results}
\vspace{-0.5cm}
\end{table}

\subsection{Image classification}\label{img_classification}
\noindent\textbf{Task and dataset.} Image classification is to identify the target correctly given the images with specific classes, which belongs to a task in CV. 
We first test the effectiveness of DiscrimLoss by performing image classification experiments on four benchmarks: MNIST~\cite{lecun2010mnist}, CIFAR-10, CIFAR-100~\cite{krizhevsky2009learning}, and Clothing1M~\cite{xiao2015learning}. 
We briefly summarize the statistics of four datasets in Table~\ref{datasets_img_classification}. 
Besides, we inject label noise into the first three datasets.
For MNIST, the noise rates are set to 0\%, 20\%, 40\%, 60\%, and 80\%. For CIFAR-10 and CIFAR-100, the noise proportions are specified as 0\%, 20\%, 40\%, and 60\%.


\noindent\textbf{Baselines.} For comparison, six categories of baselines are involved. 
(1) The model trained with vanilla cross-entropy (\textbf{CE}).
(2) Multi-network learning methods, addressing the issue of prediction error accumulation in single network learning: (i) \textbf{Co-teaching}~\cite{han2018co} maintains two networks. Each network aims to teach the other with the small-loss principle; (ii) \textbf{Co-teaching$+$}~\cite{yu2019does} comprises two networks and keeps prediction disagreement data. Small-loss data is picked and used to cross-train two DNNs; (iii) \textbf{JoCoR}~\cite{wei2020combating} trains two networks by the joint loss with co-regularization to make predictions. The objective is to reduce the diversity of the two networks during training. 
(3) Sample reweighting methods: (i) \textbf{MentorNet}~\cite{jiang2018mentornet}, which requires an extra neural network to learn a curriculum so as to alleviate overfitting on incorrect data; (ii) \textbf{Data Parameters}~\cite{saxena2019data} learns two types of learnable parameters to realize dynamic curriculum learning; (iii) 
\textbf{SuperLoss}~\cite{castells2020superloss} learns a curriculum by reweighting samples according to sample difficulty. Each sample weight can be calculated by the respective model loss. 
(4) Loss correction methods: \textbf{Forward \& Backward}~\cite{patrini2017making}, which multiplies the model prediction by a stochastic matrix $T$ or an inverse of $T$ and achieves loss correction. 
(5) Regularization-based methods: 
(i) \textbf{SLN}~\cite{chen2020noise}, which does label perturbation in SGD, provides an implicit regularization effect for training overparameterized DNNs; 
(ii) \textbf{CDR}~\cite{xia2020robust}, which divides all model parameters into critical and non-critical ones, performs different update rules for different types of parameters to hinder the memorization of noisy labels.
(6) Re-designing more robust loss methods: \textbf{CurriculumLoss}~\cite{lyu2019curriculum}, which proposes a loss to achieve curriculum sample selection. 
More details are in Appendix~\ref{appendix_img_class}.

\noindent\textbf{Implementation and Results.}
To compare with the relevant state-of-the-art (SOTA) method, for MNIST, we train a LeNet \cite{lecun2010mnist} following the experiment settings in \cite{castells2020superloss} for 20 epochs.
For CIFAR-10 and CIFAR-100, we use the WideResNet-28-10 model with the same experiment settings as those in \cite{saxena2019data} for 120 epochs.\footnote{https://github.com/apple/ml-data-parameters} 
\textcolor{black}{We utilize the SGD optimizer with batchsize 128, momentum 0.9, weight decay 5e-4, initial learning rate 0.1 for these three datasets. For CIFAR-10 and CIFAR-100, the learning rate decreased to 0.01 after 80 epochs and 0.001 after 100 epochs.}
\textcolor{black}{For Clothing1M, we train an Imagenet-pretrained ResNet-50 using the SGD optimizer with momentum 0.9, weight decay 1e-3, and batchsize 32 following the common settings in \cite{patrini2017making} and \cite{chen2020noise} for 10 epochs. The initial learning rate is 1e-3 and decreased to 1e-4 after 5 epochs.\footnote{https://github.com/chenpf1025/SLN}}

\textbf{MNIST.}
As shown in Table~\ref{MNIST-results}, 
DiscrimLoss consistently outperforms the baseline methods under different noise proportions on MNIST. In addition, with the increase of the noise level, DiscrimLoss can achieve a statistically significant gain of 0.11\%, 0.12\%, 0.24\%, 
0.62\% over the SOTA approach
, which confirms our motivation that DiscrimLoss can effectively discriminate between hard samples and incorrect samples.

\textbf{CIFAR-10 \& CIFAR-100.}
\textcolor{black}{For CIFAR-10 and CIFAR-100, we observe a similar conclusion}, as shown in Table~\ref{CIFAR10-CIFAR100-results}.
Moreover, we further explore the effectiveness of DiscrimLoss in separating hard samples from incorrect ones. Fig.~\ref{hist_9} demonstrates the distribution statistics of 
normalized losses from samples trained with cross-entropy or DiscrimLoss at 40-th, 80-th, and 120-th epochs.
We discover that DiscrimLoss can effectively distinguish incorrect samples from correct ones with an increase in training epochs by ``pushing'' 
losses of correct samples to the minimum and incorrect samples to the maximum.

\textbf{Clothing1M.}
Clothing1M is a large-scale benchmark (1$M$ training images) with real-world noise. The number of images of each class is unbalanced, and we experiment with the noisy-class-balanced sampling, following the settings in~\cite{chen2020noise}. As shown in Table~\ref{Clothin1M-results}, DiscrimLoss consistently outperforms all baselines, which indicates that DiscrimLoss can effectively discriminate among samples with various difficulties for instances within a challenging real-world noisy environment. 

\textbf{Robust loss as the inner loss.}
To demonstrate the effectiveness of DiscrimLoss on top of a robust loss, we take SuperLoss as the robust loss and place DiscrimLoss on top of it (i.e., SuperLoss-DiscrimLoss in Table~\ref{Clothin1M-results}). 
As shown in Table~\ref{Clothin1M-results}, SuperLoss-DiscrimLoss further achieves higher test accuracy compared to SuperLoss and DiscrimLoss, 
which indicates DiscrimLoss can boost robust losses effectively by appending on top of them.
\begin{table*}[htbp]
\centering
\caption{Mean absolute error (MAE) on the test set of UTKFace under different proportions of label noise.}
\begin{tabular}{|c|c|l|l|l|l|l|}
\hline
\multirow{2}{*}{Input Loss}   & \multirow{2}{*}{Method} & \multicolumn{5}{c|}{UTKFace}    \\ \cline{3-7}
&                         & \multicolumn{1}{c|}{0\%} & \multicolumn{1}{c|}{20\%} & \multicolumn{1}{c|}{40\%} & \multicolumn{1}{c|}{60\%} & \multicolumn{1}{c|}{80\%} \\ \hline
\multirow{3}{*}{MSE($l_2$)}   
& No CL                & 7.60$\pm$0.16  & 10.05$\pm$0.41  & 12.47$\pm$0.73  & 15.42$\pm$1.13  & 22.19$\pm$3.06           \\
& SuperLoss               & 7.24$\pm$0.47  & 8.35$\pm$0.17   & 9.10$\pm$0.33   & 11.74$\pm$0.14  & 13.91$\pm$0.13           \\ \cline{2-7}
& DiscrimLoss                    & \textbf{5.34$\pm$0.03}      & \textbf{6.22$\pm$0.04}       & \textbf{7.94$\pm$0.10}      & \textbf{10.24$\pm$0.20}    & \textbf{12.87$\pm$0.07}       \\ \hline 
\multirow{3}{*}{smooth-$l_1$} 
& No CL                & 6.98$\pm$0.19  & 7.40$\pm$0.18   & 8.38$\pm$0.08   & 11.62$\pm$0.08  & 17.56$\pm$0.33           \\
& SuperLoss               & 6.74$\pm$0.14  & 6.99$\pm$0.09   & 7.65$\pm$0.06   & 9.86$\pm$0.27   & 13.09$\pm$0.05           \\ \cline{2-7}
& DiscrimLoss          & \textbf{5.32$\pm$0.03}     &\textbf{5.77$\pm$0.03}   & \textbf{6.86$\pm$0.01}  & \textbf{9.53$\pm$0.21}         & \textbf{12.73$\pm$0.19}        \\ \hline
\end{tabular}
\label{utkface-table}
\end{table*}
\subsection{Image regression}\label{img_regression}
\noindent\textbf{Task and dataset.} 
To further test the generality of DiscrimLoss, we perform experiments on an image regression task (i.e., a CV regression task, aiming to predict the age given face images) on UTKFace~\cite{zhang2017age}, containing 23,705 aligned and cropped face images. We follow the data partition settings in \cite{zhang2017age} and randomly split  90\% as the training set and 10\% as the test set. Likewise, we manually add different noise levels to the dataset, with specific noise rates of 0\%, 20\%, 40\%, 60\%, 80\%, respectively.

\noindent\textbf{Implementation and Results.} We compare with two baselines: (1) \textbf{No CL} trains the model with the vanilla MSE loss or smooth-$l_1$ loss; (2) \textbf{SuperLoss} \cite{castells2020superloss}. 
We follow the experiment settings in \cite{castells2020superloss} and train a ResNet-18 model (with a single output) for 100 epochs using SGD with batchsize 128, weight decay 5e-4, and momentum 0.9. We utilize MAE as the evaluation metric and report the results based on $l_2$ loss and smooth-$l_1$ loss, which corresponds to a learning rate of 0.001 and 0.1, respectively. Threshold $k_1$ is assigned as $k_1$ = EMA.

As shown in Table~\ref{utkface-table},
the DiscrimLoss consistently outperforms both of the baselines by a large margin, regardless of the noise level.
This shows the versatility of DiscrimLoss in different task forms.
\begin{table}[t]
\setlength\tabcolsep{1.0pt}
\centering
\caption{Mean absolute error (MAE) on the test set of Digit Sum under different proportions of label noise.}
\begin{tabular}{|c|llll|}
\hline
\multirow{2}{*}{Method} & \multicolumn{4}{c|}{Digit Sum}                                                                                              \\ \cline{2-5}
& \multicolumn{1}{c|}{0\%} & \multicolumn{1}{c|}{20\%} & \multicolumn{1}{c|}{40\%} & \multicolumn{1}{c|}{60\%} \\ \hline
No CL                   & \multicolumn{1}{l|}{9.74$\pm$0.77}           & \multicolumn{1}{l|}{20.32$\pm$1.07}           & \multicolumn{1}{l|}{24.97$\pm$1.55}           & 30.42$\pm$1.15          \\
Baby Steps              & \multicolumn{1}{l|}{19.89$\pm$2.97}          & \multicolumn{1}{l|}{31.92$\pm$4.72}           & \multicolumn{1}{l|}{31.12$\pm$1.04}           & 32.63$\pm$1.74           \\ \hline 
DiscrimLoss($k_1$ = EMA) & \multicolumn{1}{l|}{9.54$\pm$1.26}           & \multicolumn{1}{l|}{18.72$\pm$1.44}           & \multicolumn{1}{l|}{23.50$\pm$1.78}           & 28.72$\pm$0.84           \\
DiscrimLoss($k_1$ = GA)    & \multicolumn{1}{l|}{9.42$\pm$1.19}           & \multicolumn{1}{l|}{\textbf{15.15$\pm$1.16}}      & \multicolumn{1}{l|}{\textbf{22.87$\pm$2.08}}  & \textbf{27.97$\pm$1.71}           \\
DiscrimLoss($k_1$ = 0.5)    & \multicolumn{1}{l|}{\textbf{6.45$\pm$0.72}}        & \multicolumn{1}{l|}{17.56$\pm$0.98}  & \multicolumn{1}{l|}{24.52$\pm$1.33}   & 29.01$\pm$1.92    \\ \hline
\end{tabular}
\label{digitsum-table}
\vspace{-0.3cm}
\end{table}
\subsection{Text sequence regression}\label{seq_regression}
\noindent\textbf{Task and dataset.}
To evaluate the effectiveness of our approach in a few-shot situation, we perform experiments on a low-resource text sequence regression task~\cite{cirik2016visualizing}, i.e., predicting the sum of digits when given a sequence of symbols of numbers, which can be regarded as an NLP task. ``Low resource'' depicts the scenario where only a few samples are available for model training. This task is evaluated on the Digit Sum dataset~\cite{cirik2016visualizing}, which is a synthetic dataset and comprises 1000, 200, and 200 samples as training, validation, and test sets. We reproduce the generation of the dataset and related baselines following the implementation by ~\cite{cirik2016visualizing},\footnote{https://github.com/volkancirik/curriculum-sstb} and evaluate the proposed method with manually corrupted labels. The noise rates are set to 0\%, 20\%, 40\%, and 60\%, respectively.
\noindent\textbf{Implementation and Results.}
We compare DiscrimLoss with the following baselines: (1) \textbf{No CL}, which trains the model with a $l_2$ loss on the noisy dataset. (2) \textbf{Baby Steps}\cite{spitkovsky2010baby}, which is a classic CL algorithm. It uses an incremental approach where groups of more complex examples are incrementally added to the training set, which is the strongest baseline in~\cite{cirik2016visualizing}. We follow \cite{cirik2016visualizing} and train an LSTM model~\cite{schmidhuber1997long} with hidden units of 256 without peephole connections and a batchsize of 512 for 100 epochs. We choose $l_2$ loss as the inner loss and MAE as the evaluation metric and utilize the AdamW optimizer with learning rate 0.1, weight decay 0. We report three sets of results corresponding to different threshold $k_1$ settings. 

Due to the extremely limited size of the dataset, the models are particularly vulnerable to overfit the noise within training set. 
However, as Table~\ref{digitsum-table} shows, 
our method outperforms both baseline methods under all proportions of label noise regardless of the initialization of $k_1$.\footnote{We use the code and parameter settings released by~\cite{cirik2016visualizing} to obtain baselines. Baby Steps is indeed worse than No CL, which is inconsistent with that reported in~\cite{cirik2016visualizing}.}
Especially on noise-free data, DiscrimLoss significantly reduces the error compared with the SOTA baseline.
This indicates the advantages of robust sample weights and auto-switching training objectives in our method, which prevent incurring the noise. 
\begin{table}[t]
\centering
\caption{Statistics of WikiHow dataset.}
\begin{tabular}{|c|c|c|c|}
\hline
\multirow{2}{*}{} & \multirow{2}{*}{\begin{tabular}[c]{@{}c@{}}Step\\ Infer.\end{tabular}} & \multirow{2}{*}{\begin{tabular}[c]{@{}c@{}}Goal\\ Infer.\end{tabular}} & \multirow{2}{*}{\begin{tabular}[c]{@{}c@{}}Step\\ Ordering\end{tabular}} \\
  &   &  & \\ \hline
Train Size & 336,851  & 166,708   & 752,516   \\
Validation Size   & 37,427  & 18,523   & 83,612    \\
Test Size  & 2,250  & 1,703   & 3,100    \\ \hline
\end{tabular}
\label{wikihow-statistics}
\end{table}

\begin{table}[t]
\centering
\caption{Test accuracy on WikiHow of different subtasks. $^*$ indicates our re-implementation.}
\begin{tabular}{|c|c|c|c|}
\hline
\multirow{2}{*}{Method} & \multirow{2}{*}{\begin{tabular}[c]{@{}c@{}}Step\\ Infer.\end{tabular}} & \multirow{2}{*}{\begin{tabular}[c]{@{}c@{}}Goal\\ Infer.\end{tabular}} & \multirow{2}{*}{\begin{tabular}[c]{@{}c@{}}Step\\ Ordering\end{tabular}} \\
\multicolumn{1}{|l|}{}  &   &  &  \\ \hline
Human  & 96.5  & 98.0  & 97.5    \\
BERT & 87.4  & 79.8  & 81.9    \\
XLNet  & 86.7  & 78.3 & 82.6  \\
GPT-2  & 83.6 & 68.6 & 80.1     \\
RoBERTa  & 88.2  & 82.0  & 83.5  \\
RoBERTa$^*$   & 88.21$\pm$0.08  & 79.53$\pm$0.23  &83.00$\pm$0.28  \\ \hline
RoBERTa (Ours)  &\textbf{89.36$\pm$0.40}   & \textbf{81.50$\pm$0.87}  &\textbf{83.66$\pm$0.20} \\ \hline
\end{tabular}
\label{appendix-wikihow-results}
\vspace{-0.3cm}
\end{table}
\begin{figure*}[tb]
\centering
\subfloat[ ]{
\includegraphics[width=0.55\columnwidth]{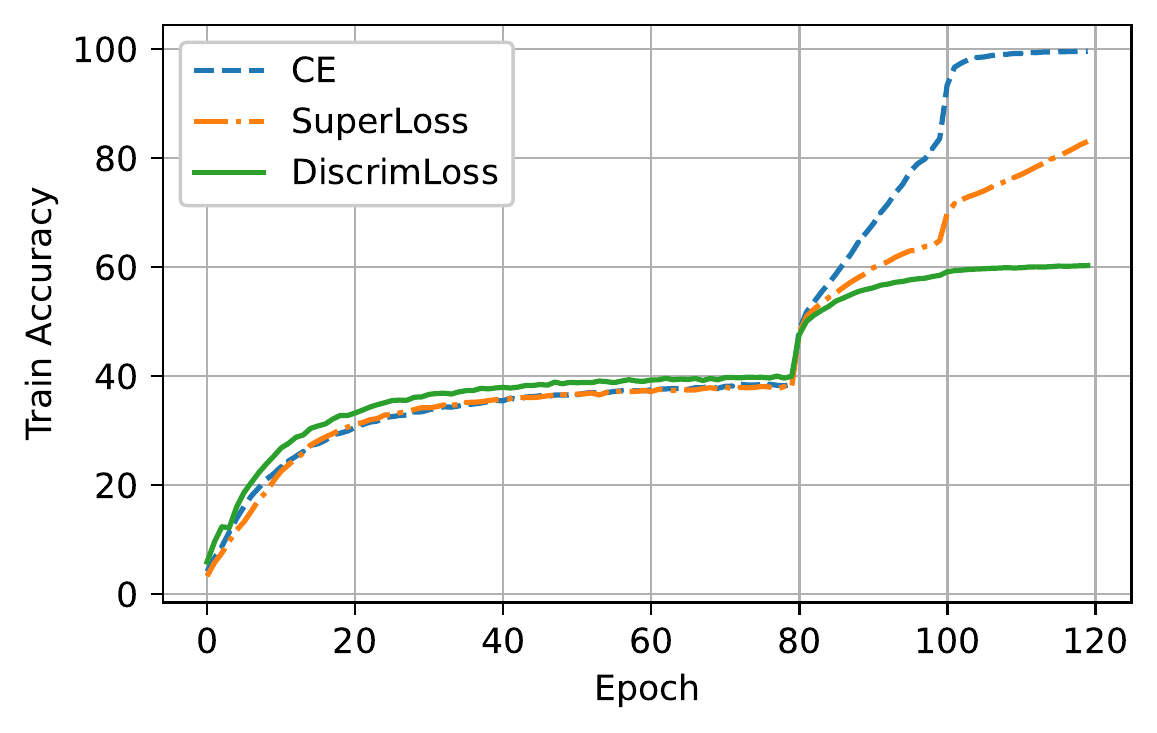}
\label{generalization_train_CIFAR100} 
}
\hfil
\subfloat[ ]{
\includegraphics[width=0.55\columnwidth]{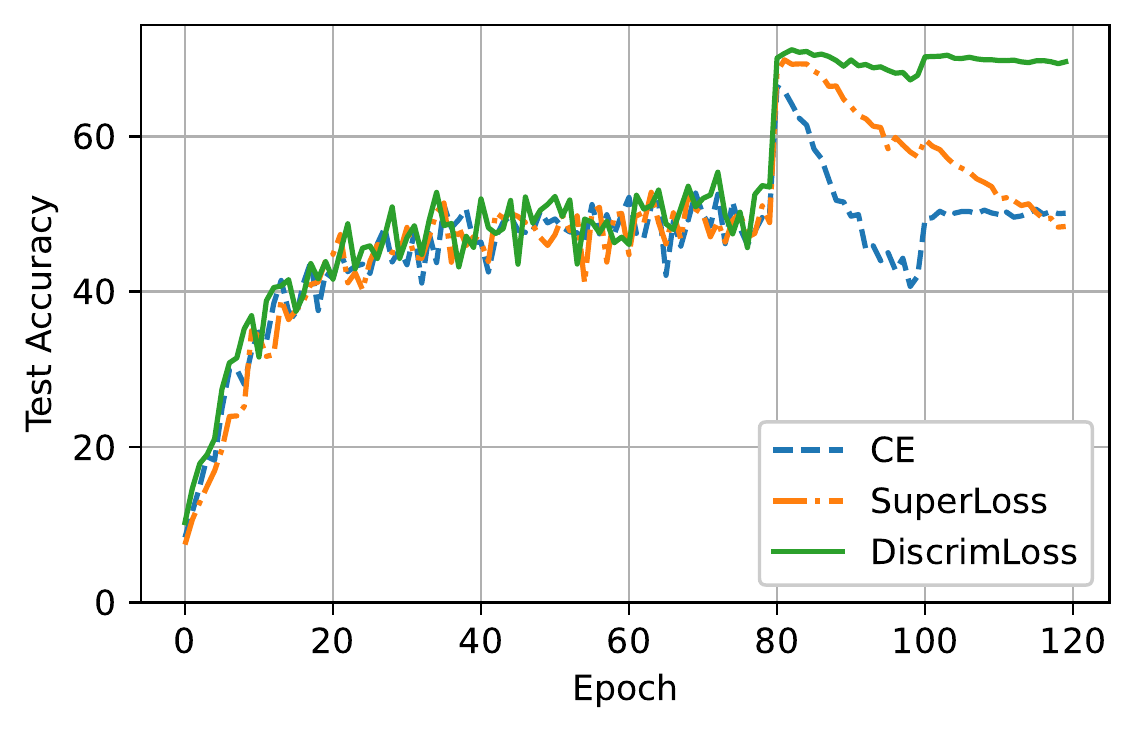}
\label{generalization_test_CIFAR100} 
}
\hfil
\subfloat[ ]{
\includegraphics[width=0.55\columnwidth]{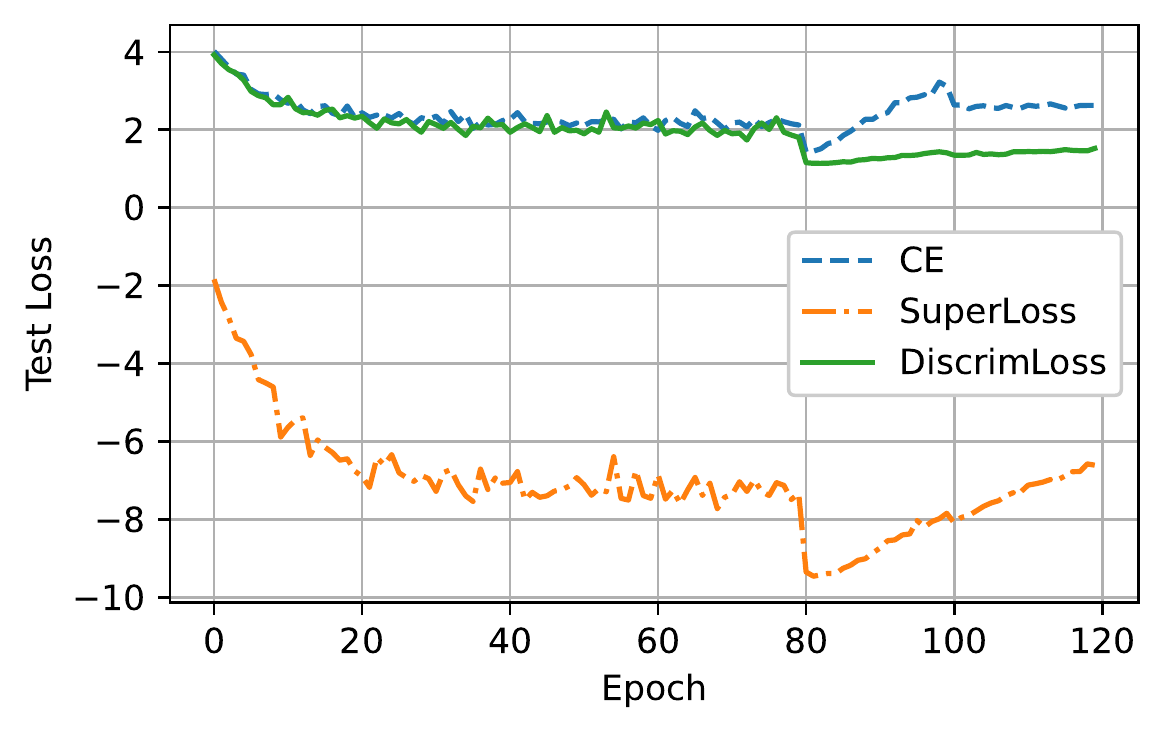}
\label{generalization_loss_CIFAR100} 
}
\caption{The generalization performance on CIFAR-100 under 40\% label noise. (a) Train accuracy. (b) Test accuracy. (c) Test loss.}
\label{generalization_total} 
\vspace{-0.3cm}
\end{figure*}
\begin{figure*}[htb]
\centering
\subfloat[ ]{
\includegraphics[width=1.2\columnwidth]{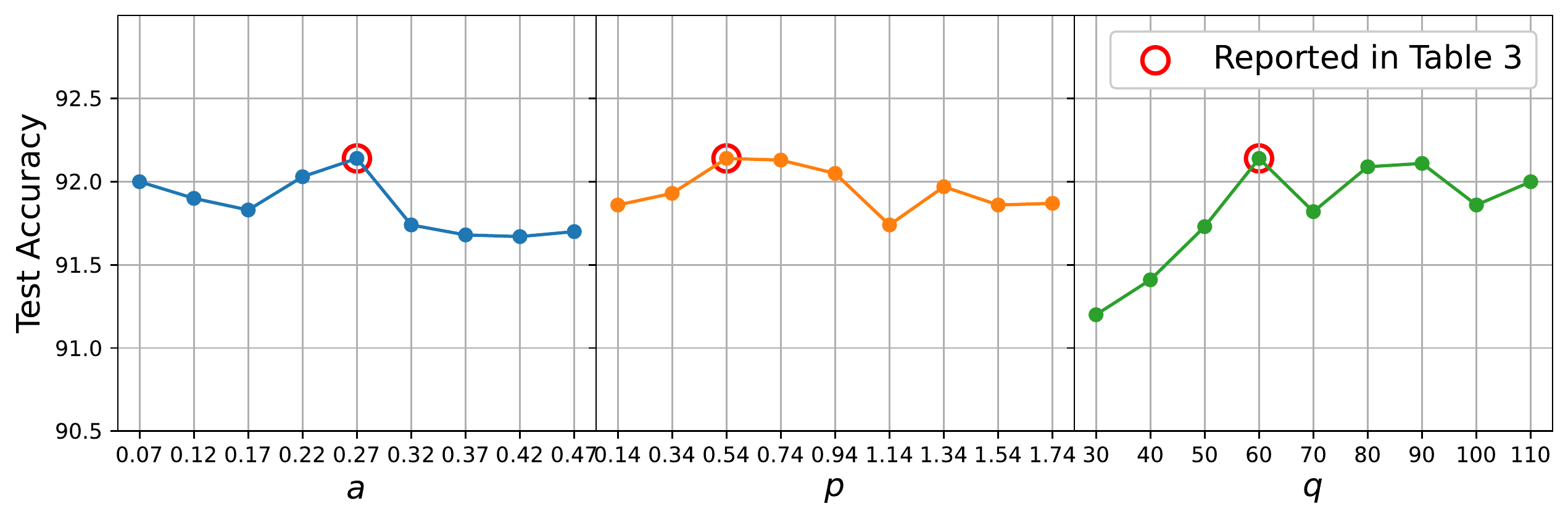}
\label{ab_abq} 
}
\hfil
\subfloat[ ]{
\includegraphics[width=0.7\columnwidth]{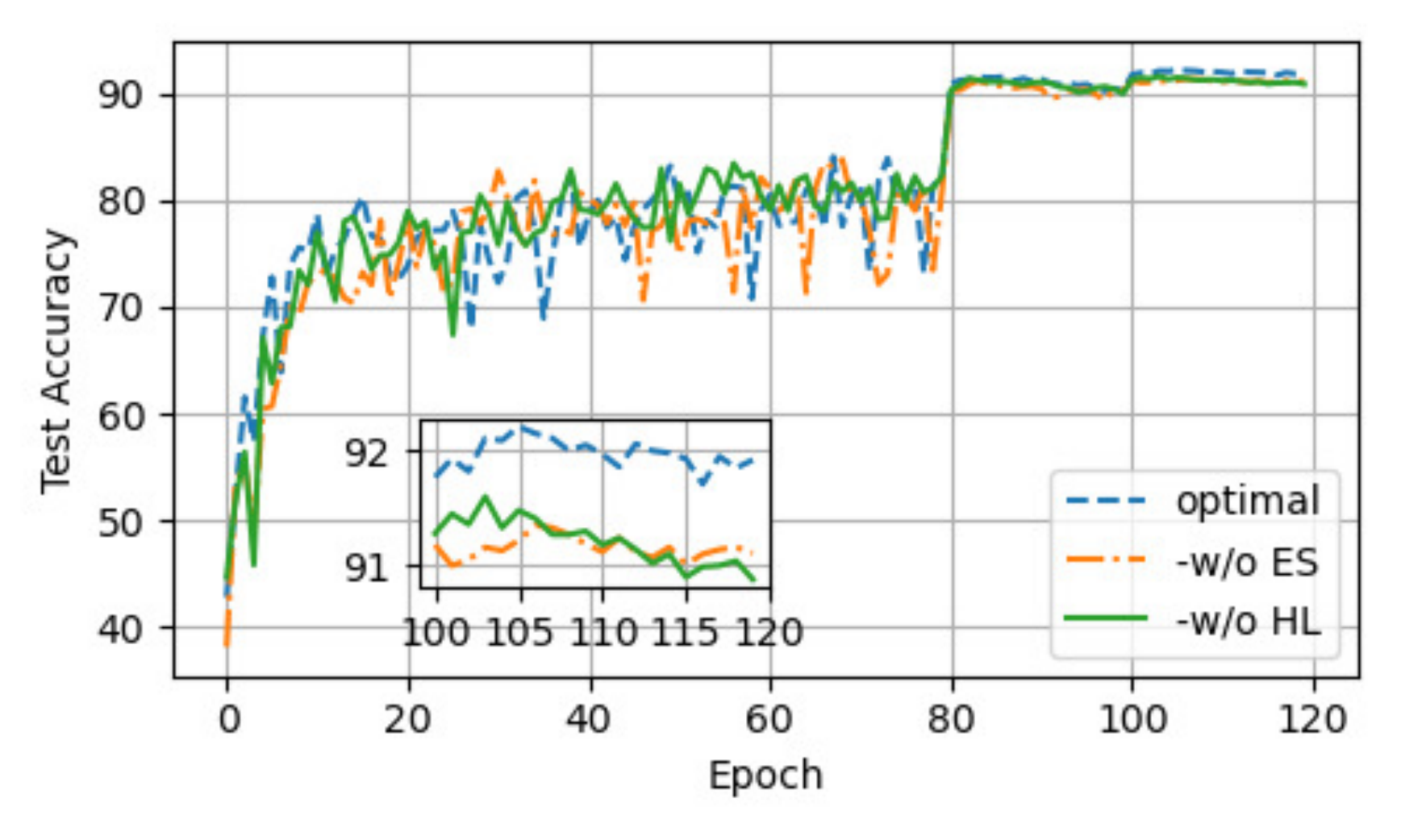}
\label{ab_wo_component} 
}
\caption{Test accuracy on CIFAR-10 under 40\% label noise. To better display the influence of changes of components on the model performance in (b), we added local amplified results of the performance curves of the last 20 epochs in the lower left part of subgraph. (a) Test accuracy w.r.t. key hyperparameters of $a$,$p$, and $q$. (b) Test accuracy after removing different key components.} 
\label{ab_total}
\vspace{-0.3cm} 
\end{figure*}
\subsection{Event relation reasoning}\label{event_rel_reasoning}
\noindent\textbf{Task and dataset.} 
To further examine the generalization ability of DiscrimLoss, we experiment on a real-world noisy NLP dataset, WikiHow~\cite{zhang2020reasoning}. 
It involves three subtasks (i.e., Step Infer., Goal Infer., and Step Ordering), aiming at relation reasoning between procedural events (i.e., GOAL-STEP relation, STEP-GOAL relation, and STEP TEMPORAL relation). 
For example, ``learn poses'' is a step in the larger goal of ``doing yoga'' and ``buy a yoga mat'' typically precedes ``learn poses'' \cite{zhang2020reasoning}. In each subtask, the training set is crawled from the wikiHow website\footnote{wikihow.com} and is automatically generated, which inevitably introduces noisy labels. The original WikiHow dataset only divides the training set and the test set. We randomly split a part of the samples from the training set as the validation set, with a ratio of 9:1. The statistics of WikiHow are shown in Table~\ref{wikihow-statistics}.

\noindent\textbf{Baselines.}
We compare the proposed method with the following methods: 
(1) \textbf{Human}~\cite{zhang2020reasoning}, which reports the human performance on these tasks. 
(2) \textbf{BERT}~\cite{JacobDevlin2018BERTPO}, which finetunes the pretrained BERT model on the training set and reports accuracy on the test set. 
(3) \textbf{XLNet}~\cite{ZhilinYang2019XLNetGA}, which is identical to BERT besides finetuning the pretrained XLNet model. 
(4) \textbf{GPT-2}~\cite{radford2019language}, which finetunes the pretrained GPT-2 model; the others are the same as BERT. 
(5) \textbf{RoBERTa}~\cite{YinhanLiu2019RoBERTaAR}, which finetunes the pretrained RoBERTa model. The remainder is similar to BERT.

\noindent\textbf{Implementation and Results.}
We reproduce the strongest baseline with the implementation by \cite{zhang2020reasoning}.\footnote{https://github.com/zharry29/wikihow-goal-step} 
We train a RoBERTa-based model using AdamW~\cite{DBLP:conf/iclr/LoshchilovH19} for three epochs (for Step Infer. subtask), two epochs (for Step Ordering subtask), and five epochs (for Goal Infer. subtask) with learning rate 5e-5, weight decay 0. Because of the large scale of the dataset, we set the batchsize as 48 and 1000 steps as an iteration. Besides, we set $e_s$ = 3, $a$ = 0.2, $p$ = 1.2, $q$ = 10, where $e_s$ and $q$ represent the number of iterations. The regularization parameter for DiscrimLoss is assigned as $\lambda$ = 1e-6, and the threshold $k_1$ is set to $k_1$ = EMA.

Table~\ref{appendix-wikihow-results} exhibits a gap of about 10\% to 20\% between model and human performance, indicating the existence of massive hard instances. In addition, compared to RoBERTa$^*$ optimized using cross-entropy loss, DiscrimLoss significantly improves the model performance by 1.15\%, 1.97\%, and 0.66\% on subtasks Step Infer., Goal Infer., and Step Ordering, respectively.



\subsection{Analysis on model generalization improvement}\label{gen_performance_ana}

We further explore the effectiveness of DiscrimLoss on improving model generalization. We analyze a manually corrupted CIFAR-100 dataset, with 40\% manually injected noisy labels, and compare with two baselines CE and SuperLoss.

As illustrated in Fig.~\ref{generalization_total}~\subref{generalization_train_CIFAR100}, the accuracy of DiscrimLoss-optimized model on the manually corrupted training set is 60.38\%, close to the theoretical upperbound 60\%. However, SuperLoss and CE obtain 83.03\% and 99.58\% accuracy, respectively, indicating apparent overfitting of the noise. 
Correspondingly, as demonstrated in Fig.~\ref{generalization_total}~\subref{generalization_test_CIFAR100}, the test accuracy of DiscrimLoss level off after about 80 epochs, rather than a significant downward trend like other methods, confirming a better generalization performance of DiscrimLoss.
We get a similar conclusion in Fig.~\ref{generalization_total}~\subref{generalization_loss_CIFAR100}, where after about 80 epochs, 
the test loss of DiscrimLoss tends to convergence, keeping stabilized. On the contrary, the other methods show a notable increase in test loss manifesting obvious overfitting. 

\subsection{Ablation study}\label{ab_study}

We explore the specific influence of key hyperparameters and components of DiscrimLoss. All experiments are conducted on the test set of CIFAR-10 under 40\% noisy labels.

The DiscrimLoss contains three crucial hyperparameters, i.e., switching speed $a$, switching threshold $p$, and switching moment $q$, for controlling the switching of different training stages, and seperating easy, hard and incorrect samples.
We study their effects by changing the value of one hyperparameter at a time, meanwhile keeping the others fixed. The default value of hyperparameters are obtained via optimization, with $a$ = 0.27, $p$ = 0.54, and $q$ = 60. 
Fig.~\ref{ab_total}~\subref{ab_abq} shows no matter altering $a$, $p$, or $q$ can lead to good performance, indicating model performance is insensitive to their selection, despite introducing several hyperparameters. Moreover, diverse noise levels on a dataset usually correspond to the same set of optimal hyperparameters in Table~\ref{optimal-hyper-with-noise-rate} (e.g., CIFAR-10 under 20\%, 40\%, 60\% label noise), which further support our claims. 
Fig.~\ref{ab_total}~\subref{ab_wo_component} shows the model performance after removing the ES or HL component. We find that removing either can lead to considerable performance degradation, demonstrating their necessity and importance in DiscrimLoss. Component ES and HL enable a stable estimation of sample weight and better distinguish easy, hard, and incorrect samples compared with existing studies.




\section{Related work}
\textbf{Curriculum learning.}
As a generic learning paradigm, CL benefits the learning process by presenting training examples in a meaningful order and has been widely applied in NLP~\cite{huang2019self,xu2020curriculum,zhao2020reinforced},  CV~\cite{hacohen2019power,soviany2020image}, and many other fields~\cite{guo2020breaking,wang2020curriculum,zheng2020unsupervised}. Early works of CL heavily rely on the quality of heuristical prior knowledge to select samples while ignoring feedback from the learner, which triggers a discrepancy between the fixed curriculum and the model. To address this shortcoming, \cite{kumar2010self} propose self-paced learning (SPL), which implements sample selection with a weight variable and no longer requires human intervention.
The weight variable and model parameters are optimized alternatively. 
Recently, several studies have been proposed to further refine weighting strategies of SPL \cite{jiang2015self,jiang2018mentornet,saxena2019data,zheng2020unsupervised}. However, they mainly depend on the auxiliary model framework or are limited by specific tasks. Like SPL, DiscrimLoss is also a method of learning a curriculum dynamically without the limitation of inconsistency. Moreover, as an all-purpose CL-based method, DiscrimLoss is suitable for many tasks without modifying model architecture.


\textbf{Learning on noisy samples.}
To alleviate the influence of incorrect samples, most previous works focus on devising sample reweighting strategy to suppress the weight of incorrect samples
\cite{jiang2018mentornet,ren2018learning,lyu2019curriculum}. 
For example, Co-teaching~\cite{han2018co} and Co-teaching$+$~\cite{yu2019does} cross-train two DNNs, which have different learning abilities, to filter the noise.
DivideMix \cite{li2020dividemix} models the per-sample loss distribution with a mixture model for correct and incorrect samples' division. \cite{lyu2019curriculum} propose CurriculumLoss, 
which can be deemed as a curriculum sample selection strategy.
SuperLoss \cite{castells2020superloss} learns a curriculum by reweighting samples according to sample difficulty. 
On the other hand, both hard samples and incorrect samples are pervasive but challenging for the model to tell one from the other. However, previous studies simply model incorrect samples as general hard ones, which is not strictly distinguished from the concept: hard samples. 
In this paper, DiscrimLoss tries to discriminate between hard samples and incorrect samples as much as possible, which benefits the model to learn hard examples but avoid memorizing from incorrect ones.

\section{Conclusion}\label{Conclusion}
Aiming at the problem of ubiquitous noise, we propose DiscrimLoss for separating hard samples and incorrect samples. Our method employs two-stage training objectives with auto-switching to concurrently guarantee the model performance as well as generalization. Moreover, from the perspective of prediction confidence enhancement, we devise efficient components into DiscrimLoss and generate more stable sample weights, which makes DiscrimLoss outperforms other baselines even in noise-free scenarios. For future work, we are interested in adapting DiscrimLoss to more domains with more datasets. The code will be released upon publication.

\section*{Acknowledgements}   
This work was supported by the Technological Innovation ``2030 Megaproject'' - New Generation Artificial Intelligence of China (Grant No. 2020AAA0106501), the National Natural Science Foundation of China (Grant No. 62176079, 61976073).    

\appendices
\section{Optimal hyperparameters settings}\label{appendix_hypersetting}
We report the optimal settings of hyperparameters unique to our method in Table~\ref{optimal-hyper-with-noise-rate} and Table~\ref{optimal-hyper}.

\section{Baselines in the image classification task}\label{appendix_img_class}
\begin{table*}[]
\centering
\caption{\textcolor{black}{Optimal hyperparameters settings on MNIST, CIFAR-10, CIFAR-100, UTKFace, and Digit Sum under different proportions of label noise, where UTKFace (1) (/UTKFace (2)) denotes the setting taking the smooth-$l_1$ loss (/$l_2$ loss) as the inner loss.}}
\begin{tabular}{|c|ccccc|}
\hline
\multirow{2}{*}{Dataset}                                               & \multicolumn{5}{c|}{Optimal Hyperparameters Settings}                                                                                                                                                          \\ \cline{2-6} 
& \multicolumn{1}{c|}{0\%}                    & \multicolumn{1}{c|}{20\%}                   & \multicolumn{1}{c|}{40\%}                   & \multicolumn{1}{c|}{60\%}                   & 80\%                   \\ \hline
\multirow{3}{*}{MNIST}                                                 & \multicolumn{1}{c|}{$e_s$ = 2, $a$ = 0.35,} & \multicolumn{1}{c|}{$e_s$ = 2, $a$ = 0.50,} & \multicolumn{1}{c|}{$e_s$ = 2, $a$ = 0.10,} & \multicolumn{1}{c|}{$e_s$ = 2, $a$ = 0.10,} & $e_s$ = 2, $a$ = 0.12, \\
& \multicolumn{1}{c|}{$p$ = 1.56, $q$ = 12,}  & \multicolumn{1}{c|}{$p$ = 1.05, $q$ = 2,}   & \multicolumn{1}{c|}{$p$ = 0.97, $q$ = 18,}  & \multicolumn{1}{c|}{$p$ = 0.61, $q$ = 16,}  & $p$ = 1.20, $q$ = 14,  \\
& \multicolumn{1}{c|}{$\lambda$ = 0}          & \multicolumn{1}{c|}{$\lambda$ = 8e-3}       & \multicolumn{1}{c|}{$\lambda$ = 0}          & \multicolumn{1}{c|}{$\lambda$ = 0}          & $\lambda$ = 0.09       \\ \hline
\multirow{3}{*}{CIFAR-10}                                              & \multicolumn{1}{c|}{$e_s$ = 4, $a$ = 0.25,} & \multicolumn{1}{c|}{$e_s$ = 3, $a$ = 0.27,} & \multicolumn{1}{c|}{$e_s$ = 3, $a$ = 0.27,} & \multicolumn{1}{c|}{$e_s$ = 3, $a$ = 0.27,} & \multirow{3}{*}{-}     \\
& \multicolumn{1}{c|}{$p$ = 1.92, $q$ = 81,}  & \multicolumn{1}{c|}{$p$ = 0.54, $q$ = 60,}  & \multicolumn{1}{c|}{$p$ = 0.54, $q$ = 60,}  & \multicolumn{1}{c|}{$p$ = 0.54, $q$ = 60,}  &                        \\
& \multicolumn{1}{c|}{$\lambda$ = 0}          & \multicolumn{1}{c|}{$\lambda$ = 0}          & \multicolumn{1}{c|}{$\lambda$ = 0}          & \multicolumn{1}{c|}{$\lambda$ = 0}          &                        \\ \hline
\multirow{3}{*}{CIFAR-100}                                             & \multicolumn{1}{c|}{$e_s$ = 4, $a$ = 0.25,} & \multicolumn{1}{c|}{$e_s$ = 4, $a$ = 0.25,} & \multicolumn{1}{c|}{$e_s$ = 3, $a$ = 0.27,} & \multicolumn{1}{c|}{$e_s$ = 2, $a$ = 0.37,} & \multirow{3}{*}{-}     \\
& \multicolumn{1}{c|}{$p$ = 1.92, $q$ = 81,}  & \multicolumn{1}{c|}{$p$ = 1.92, $q$ = 81,}  & \multicolumn{1}{c|}{$p$ = 0.54, $q$ = 60,}  & \multicolumn{1}{c|}{$p$ = 2.25, $q$ = 75,}  &                        \\
& \multicolumn{1}{c|}{$\lambda$ = 0}          & \multicolumn{1}{c|}{$\lambda$ = 0}          & \multicolumn{1}{c|}{$\lambda$ = 0}          & \multicolumn{1}{c|}{$\lambda$ = 0}          &                        \\ \hline
\multirow{3}{*}{\begin{tabular}[c]{@{}c@{}}UTKFace\\ (1)\end{tabular}} & \multicolumn{1}{c|}{$e_s$ = 3, $a$ = 0.42,} & \multicolumn{1}{c|}{$e_s$ = 2, $a$ = 0.46,} & \multicolumn{1}{c|}{$e_s$ = 2, $a$ = 0.46,} & \multicolumn{1}{c|}{$e_s$ = 4, $a$ = 0.25,} & $e_s$ = 3, $a$ = 0.42, \\
& \multicolumn{1}{c|}{$p$ = 1.40, $q$ = 41,}  & \multicolumn{1}{c|}{$p$ = 1.25, $q$ = 50,}  & \multicolumn{1}{c|}{$p$ = 1.25, $q$ = 50,}  & \multicolumn{1}{c|}{$p$ = 1.92, $q$ = 28,}  & $p$ = 1.40, $q$ = 41,  \\
& \multicolumn{1}{c|}{$\lambda$ = 0.09}       & \multicolumn{1}{c|}{$\lambda$ = 0.01}       & \multicolumn{1}{c|}{$\lambda$ = 0.01}       & \multicolumn{1}{c|}{$\lambda$ = 3e-6}       & $\lambda$ = 0.09       \\ \hline
\multirow{3}{*}{\begin{tabular}[c]{@{}c@{}}UTKFace\\ (2)\end{tabular}} & \multicolumn{1}{c|}{$e_s$ = 3, $a$ = 0.42,} & \multicolumn{1}{c|}{$e_s$ = 2, $a$ = 0.46,} & \multicolumn{1}{c|}{$e_s$ = 2, $a$ = 0.46,} & \multicolumn{1}{c|}{$e_s$ = 2, $a$ = 0.46,} & $e_s$ = 3, $a$ = 0.42, \\
& \multicolumn{1}{c|}{$p$ = 1.40, $q$ = 41,}  & \multicolumn{1}{c|}{$p$ = 1.25, $q$ = 50,}  & \multicolumn{1}{c|}{$p$ = 1.25, $q$ = 50,}  & \multicolumn{1}{c|}{$p$ = 1.25, $q$ = 50,}  & $p$ = 1.40, $q$ = 41,  \\
& \multicolumn{1}{c|}{$\lambda$ = 0.09}       & \multicolumn{1}{c|}{$\lambda$ = 0.01}       & \multicolumn{1}{c|}{$\lambda$ = 0.01}       & \multicolumn{1}{c|}{$\lambda$ = 0.01}       & $\lambda$ = 0.09       \\ \hline
\multirow{3}{*}{Digit Sum}                                             & \multicolumn{1}{c|}{$e_s$ = 3, $a$ = 1.51,} & \multicolumn{1}{c|}{$e_s$ = 3, $a$ = 0.48,} & \multicolumn{1}{c|}{$e_s$ = 3, $a$ = 1.18,} & \multicolumn{1}{c|}{$e_s$ = 3, $a$ = 1.09,} & \multirow{3}{*}{-}     \\
& \multicolumn{1}{c|}{$p$ = 3.17, $q$ = 67,}  & \multicolumn{1}{c|}{$p$ = 3.03, $q$ = 57,}  & \multicolumn{1}{c|}{$p$ = 0.23, $q$ = 54,}  & \multicolumn{1}{c|}{$p$ = 1.37, $q$ = 75,}  &                        \\
& \multicolumn{1}{c|}{$\lambda$ = 9e-8}       & \multicolumn{1}{c|}{$\lambda$ = 0.550}      & \multicolumn{1}{c|}{$\lambda$ = 0.105}      & \multicolumn{1}{c|}{$\lambda$ = 0.145}      &                        \\ \hline
\end{tabular}
\label{optimal-hyper-with-noise-rate}
\vspace{-0.5cm}
\end{table*}
\textcolor{black}{Similar to baselines selection in~\cite{xia2020robust}, we do not compare with some SOTA methods like SELF~\cite{nguyen2019self} and DivideMix~\cite{li2020dividemix} as baselines because of the following reasons.
(1) Their proposed methods are aggregations of multiple techniques while we only focus on one. Therefore the comparison is not fair. 
(2) We focus on essentially solving the problem of memorization of noisy labels, i.e., how to automatically separating hard and incorrect samples as much as possible, but not on boosting the classification performance.}

\begin{table}[]
\centering
\caption{Optimal hyperparameters settings on Clothing1M and WikiHow.}
\begin{tabular}{|c|c|}
\hline
Dataset                     & Optimal Hyperparameters Settings \\ \hline
\multirow{6}{*}{Clothing1M} & $e_s$ = 3, $a$ = 0.26, $p$ = 0.68  \\
& $q$ = 5, $\lambda$ = 0           \\
& (for DiscrimLoss in Table~\ref{Clothin1M-results}) \\ \cline{2-2}
& $e_s$ = 3, $a$ = 0.16, $p$ = 1.07  \\
& $q$ = 3, $\lambda$ = 0           \\ 
& (for SuperLoss-DiscrimLoss in Table~\ref{Clothin1M-results}) \\ \hline
\multirow{2}{*}{WikiHow}    & $e_s$ = 3, $a$ = 0.2, $p$ = 1.2  \\
& $q$ = 10, $\lambda$ = 1e-6       \\ \hline
\end{tabular}
\label{optimal-hyper}
\vspace{-0.5cm}
\end{table}
\bibliographystyle{IEEEtran}
\bibliography{reference}
\vspace{-33pt}
\begin{IEEEbiographynophoto}{Tingting Wu} received the master’s degree from University of Science and Technology of China. She is working towards a Ph.D. degree in the School of Computer Science and Technology at Harbin Institute of Technology. Her research interests include machine learning, label-noise learning, and natural language processing (NLP) applications.
\end{IEEEbiographynophoto}
\vspace{-33pt}
\begin{IEEEbiographynophoto}{Xiao Ding} received the Ph.D. degree from the School of Computer Science and Technology, Harbin Institute of Technology, where he is currently an Associate Researcher. His research interests include natural language processing, text mining, social computing, and commonsense inference.
\end{IEEEbiographynophoto}
\vspace{-33pt}
\begin{IEEEbiographynophoto}{Hao Zhang} is an Associate Researcher in the Computer Science Department at Harbin Institute of Technology (HIT), China. He received the BS and Ph.D. degrees from University of Science and Technology of China in 2014. His research interests include parallel computing, federated learning, pervasive computing, and deep learning application.
\end{IEEEbiographynophoto}
\vspace{-33pt}
\begin{IEEEbiographynophoto}{Jinglong Gao} received his bachelor's degree in Computer Science and Technology from Harbin Institute of Technology, Harbin, China, in 2020. He is currently pursuing a Ph.D. degree in Computer Science and Technology at Harbin Institute of Technology, Harbin, China. His research interests include natural language processing, causality identification, and event-centric reasoning.
\end{IEEEbiographynophoto}
\vspace{-33pt}
\begin{IEEEbiographynophoto}{Li Du} received a master's degree in Statistics from Fudan University, China, in July 2017. Since 2018, he has been working toward a Ph.D. degree in the Department of Computer Science, Harbin Institute of Technology. His current research interests include natural language processing, commonsense reasoning, and pretrained language models.
\end{IEEEbiographynophoto}
\vspace{-33pt}
\begin{IEEEbiographynophoto}{Bing Qin} received a Ph.D. degree from the Department of Computer Science, Harbin Institute of Technology, China, in 2005. She is currently a full professor in the Department of Computer Science, and the director of the Research Center for Social Computing and Information Retrieval (HIT-SCIR), Harbin Institute of Technology. Her research interests include natural language processing, information extraction, document-level discourse analysis, and sentiment analysis.
\end{IEEEbiographynophoto}
\vspace{-33pt}
\begin{IEEEbiographynophoto}{Ting Liu} received a Ph.D. degree from the Department of Computer Science, Harbin Institute of Technology, China, in 1998. He is currently a full professor in the Department of Computer Science, and the director of the Research Center for Social Computing and Information Retrieval (HIT-SCIR), Harbin Institute of Technology. His research interests include information retrieval, natural language processing, and social media analysis.
\end{IEEEbiographynophoto}

\vfill

\end{document}